\documentclass[11pt]{article}
\usepackage[T1]{fontenc}
\usepackage[utf8]{inputenc}
\usepackage{lmodern}
\usepackage[margin=0.94in]{geometry}
\usepackage{amsmath,amssymb,amsthm,mathtools}
\usepackage{booktabs,array,tabularx,longtable}
\usepackage{graphicx}
\usepackage{placeins}
\usepackage{flafter}
\usepackage{natbib}
\usepackage{microtype}
\usepackage{algorithm}
\usepackage{enumitem}
\usepackage{listings}
\usepackage{tikz}
\usetikzlibrary{positioning,calc,fit}
\usepackage[dvipsnames]{xcolor}
\usepackage[hidelinks]{hyperref}
\hypersetup{pdftitle={Training Autoregressive Language Models on Rounded, Bounded, and Censored Numerical Labels via Exact Likelihood},pdfsubject={Exact boundary marginalization for rounded, bounded, and censored numerical labels},pdfkeywords={numerical language models, censored supervision, interval likelihood, CRPS}}
\usepackage{fancyhdr}
\setlist{nosep,leftmargin=*}
\newtheorem{proposition}{Proposition}

\newcommand{\E}{\mathbb{E}}

\newcommand{\ind}{\mathbf{1}}
\newcommand{\KL}{\operatorname{KL}}
\newcommand{\LSE}{\operatorname{LSE}}

\newcommand{\Cov}{\operatorname{Cov}}

\newcommand{\cV}{\mathcal{V}}
\newcommand{\loss}{\mathcal{L}}
\newcommand{\logmass}{\operatorname{LogMass}}
\newcommand{\paperfigure}[2]{\IfFileExists{#1}{\includegraphics[width=#2]{#1}}{\fbox{\parbox{.85\linewidth}{\small Figure asset: \texttt{\detokenize{#1}}. Copy the supplied manuscript\_figures directory.}}}}
\newcolumntype{Y}{>{\raggedright\arraybackslash}X}
\title{\textbf{\LARGE Training Autoregressive Language Models on Rounded, Bounded, and Censored Numerical Labels via Exact Likelihood}}

\author{
Jingquan Yan$^{1}$, Yuwei Miao$^{1}$, Yang Xie$^{2}$, Junzhou Huang$^{1}$\\[4pt]
{\small $^{1}$University of Texas at Arlington}\\
{\small $^{2}$Washington University in St. Louis}\\[4pt]
{\small \texttt{jingquan.yan@uta.edu, yuwei.miao@uta.edu}}\\
{\small \texttt{xieyang@wustl.edu, jzhuang@uta.edu}}
}

\date{}
\begin{document}
\maketitle
\begin{abstract}
Numerical labels often specify a set of possible values: a measurement is rounded, a laboratory test reports only a bound, or a program is stopped at a timeout. We introduce boundary marginalization, the first method that trains an autoregressive language model on the exact likelihood of such observations through its existing digit tokens. A fixed-width, order-preserving numeral representation groups consecutive grid values under shared digit prefixes. An interval is a union of such groups, so its exact probability and gradient follow from at most two boundary paths, the successive digit prefixes of its endpoints, each of length logarithmic in the number of grid values. The resulting objective unifies precise, rounded, bounded, and censored supervision and trains digit decisions at the resolution supplied by each observation. The same computation supports exact distributional queries and unbiased gradients for the continuous ranked probability score on exact labels. Experiments show recovery of a conditional distribution from rounded and censored reports alone, improved likelihood on naturally censored molecular measurements, and more accurate code latency distributions on two backbones. Exact interval updates remain tractable as the numerical support grows to one million values.

\end{abstract}

\section{Introduction}
Numerical observations often report a set of compatible values, which we call a \emph{numerical event}. A measurement rounded to one decimal place and reported as $1.2$ represents the interval $[1.15,1.25)$. A drug-safety test may report only that a compound's half-maximal inhibitory concentration (IC50), the concentration at which it inhibits its target by 50\%, exceeds the highest concentration tested. A program unfinished at a timeout $c$ has a completion time greater than $c$. A report that provides only a bound is a \emph{censored observation}.

Such reports are common and informative. In the cardiac-safety measurements we study, about a third of the labels are bounds (Section~\ref{sec:experiments}), and Svensson et al.\ report that a third or more of the labels are censored in several drug-discovery datasets \citep{svensson2025}. Cross-entropy training requires a target numeral, so standard practice either discards these reports or replaces each one with a representative numeral, which the model then learns to reproduce. Training on the total probability of the compatible values uses each report exactly as observed.

Autoregressive language models (LMs) already define probability distributions over numerical values through their digit tokens. LLMTime interprets digit sequences as a hierarchical numerical distribution, and decoding-based regression develops numerical decoder representations and likelihood training \citep{gruver2023,song2025}. The probability of an individual numeral is directly available, but a reported interval can contain thousands of numerals. How can an LM evaluate and differentiate their total probability without enumerating them?

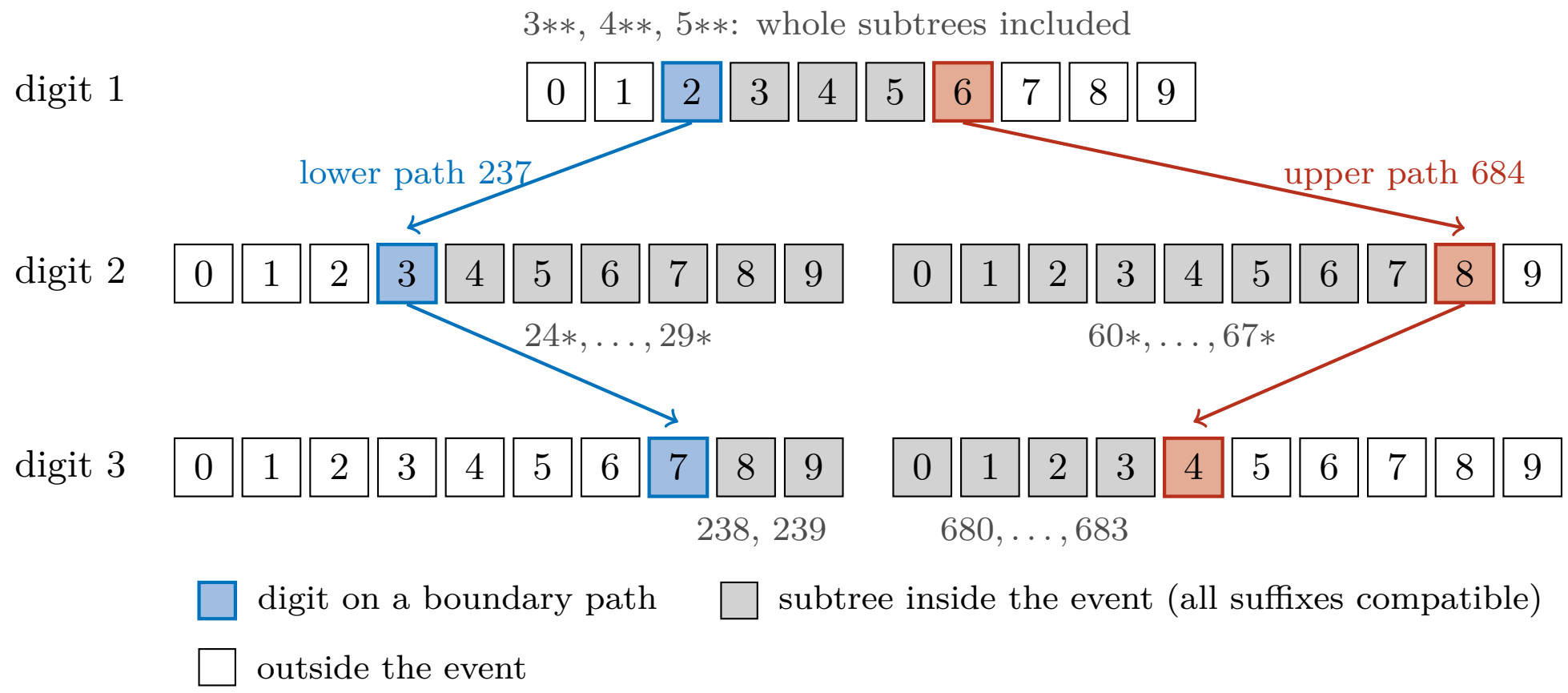
\begin{figure}[t]
\centering
\begin{tikzpicture}[x=0.56cm,y=1cm,
  box/.style={draw,minimum width=0.48cm,minimum height=0.48cm,inner sep=0pt,font=\footnotesize,fill=white},
  lower/.style={box,fill=RoyalBlue!35,draw=RoyalBlue,thick},
  upper/.style={box,fill=BrickRed!35,draw=BrickRed,thick},
  inc/.style={box,fill=gray!35},
  lab/.style={font=\footnotesize,anchor=east},
  grp/.style={font=\scriptsize,text=black!70}]
\node[lab] at (-10.6,0) {digit 1};
\foreach \d in {0,1,7,8,9} {\node[box] (a\d) at (\d-4.5,0) {\d};}
\node[lower] (a2) at (-2.5,0) {2};
\foreach \d in {3,4,5} {\node[inc] (a\d) at (\d-4.5,0) {\d};}
\node[upper] (a6) at (1.5,0) {6};
\node[grp,above=1pt of a4] {$3{*}{*},\,4{*}{*},\,5{*}{*}$: whole subtrees included};
\node[lab] at (-10.6,-1.5) {digit 2};
\foreach \d in {0,1,2} {\node[box] (b\d) at (\d-9.7,-1.5) {\d};}
\node[lower] (b3) at (-6.7,-1.5) {3};
\foreach \d in {4,5,6,7,8,9} {\node[inc] (b\d) at (\d-9.7,-1.5) {\d};}
\node[grp,below=1pt of b6.south west,anchor=north west,xshift=-0.6cm] {$24{*},\ldots,29{*}$};
\foreach \d in {0,1,2,3,4,5,6,7} {\node[inc] (c\d) at (\d+0.9,-1.5) {\d};}
\node[upper] (c8) at (8.9,-1.5) {8};
\node[box] (c9) at (9.9,-1.5) {9};
\node[grp,below=1pt of c3.south west,anchor=north west,xshift=-0.2cm] {$60{*},\ldots,67{*}$};
\node[lab] at (-10.6,-3.1) {digit 3};
\foreach \d in {0,1,2,3,4,5,6} {\node[box] (d\d) at (\d-9.7,-3.1) {\d};}
\node[lower] (d7) at (-2.7,-3.1) {7};
\foreach \d in {8,9} {\node[inc] (d\d) at (\d-9.7,-3.1) {\d};}
\node[grp,below=1pt of d8.south west,anchor=north west,xshift=-0.3cm] {$238,\,239$};
\foreach \d in {0,1,2,3} {\node[inc] (e\d) at (\d+0.9,-3.1) {\d};}
\node[upper] (e4) at (4.9,-3.1) {4};
\foreach \d in {5,6,7,8,9} {\node[box] (e\d) at (\d+0.9,-3.1) {\d};}
\node[grp,below=1pt of e1.south west,anchor=north west,xshift=-0.3cm] {$680,\ldots,683$};
\draw[RoyalBlue,thick,->] (a2.south) -- ($(b3.north)+(0,0.12)$) node[midway,left,font=\scriptsize,text=RoyalBlue] {lower path $237$};
\draw[RoyalBlue,thick,->] (b3.south) -- ($(d7.north)+(0,0.12)$);
\draw[BrickRed,thick,->] (a6.south) -- ($(c8.north)+(0,0.12)$) node[midway,right,font=\scriptsize,text=BrickRed] {\phantom{8pt}upper path $684$};
\draw[BrickRed,thick,->] (c8.south) -- ($(e4.north)+(0,0.12)$);
\node[lower,minimum width=0.3cm,minimum height=0.3cm] (l1) at (-9.5,-4.2) {};
\node[right=1pt of l1,font=\scriptsize] {digit on a boundary path};
\node[inc,minimum width=0.3cm,minimum height=0.3cm] (l2) at (-1.8,-4.2) {};
\node[right=1pt of l2,font=\scriptsize] {subtree inside the event (all suffixes compatible)};
\node[box,minimum width=0.3cm,minimum height=0.3cm] (l3) at (-9.5,-4.75) {};
\node[right=1pt of l3,font=\scriptsize] {outside the event};
\end{tikzpicture}
\caption{Boundary marginalization for $237\leq Y\leq684$ on the three-digit grid $v_k=k$, where $*$ denotes an arbitrary digit. Every numeral in the event either is a boundary numeral or first departs from a boundary path at one digit and then continues with an arbitrary suffix. Since the suffix probabilities of a subtree sum to one, the event probability is a sum of prefix masses times sibling digit probabilities, all evaluated at prefixes on the two boundary paths (colored). The model is never evaluated inside a shaded subtree, so the cost is independent of the $448$ values in the event.}
\label{fig:tree}
\end{figure}

Numerical order makes this possible. In a fixed-width numeral written with the most significant digit first, every value below a threshold has a unique first digit at which it becomes smaller. After that digit, all suffixes are allowed, and their conditional probabilities sum to one. Numerals sharing an initial digit sequence form a \emph{prefix subtree}. Its total probability is just the probability of that prefix. To evaluate a threshold, we need only the next-digit distributions along its \emph{boundary path}: the successive prefixes of the threshold numeral. Teacher forcing supplies these known prefixes as inputs and reads their next-digit probabilities in one forward pass. We call this computation \emph{boundary marginalization}: evaluating and differentiating the exact probability of a numerical interval from at most two teacher-forced boundary paths (Figure~\ref{fig:tree}). For $M$ grid values in base $b$, the scalar arithmetic is $O(b\log_b M)$, independent of the number of compatible values.

A single objective, the likelihood of the observed event, therefore handles precise labels, intervals, and bounds. A singleton (one grid value) recovers sequence cross-entropy (CE), the log score of the complete numeral under the same decoder. An interval supervises its total probability: only prefixes whose subtrees lie partly inside and partly outside the interval receive a direct gradient, while subtrees lying entirely inside it are summed out and need no target digit.

The same computation gives cumulative distribution function (CDF) values for continuous ranked probability score (CRPS) training and exact distributional queries. CRPS integrates the squared difference between the predicted CDF and the step function at the observed value, so it scores predictions by numerical distance: a prediction near the observed value incurs a small loss even when its digits differ. It is a strictly proper score: its expected loss is uniquely minimized by the true distribution. Its training gradient is unbiased when estimated from a few thresholds sampled independently of the model.

\paragraph{Contributions.}
We introduce boundary marginalization, the first method that trains a pretrained language model on the exact likelihood of rounded, bounded, and censored numerical observations through its existing digit tokens, with no auxiliary output head, imputed target, or sampled rollout.
\begin{enumerate}[label=(\roman*), leftmargin=2.5em, itemsep=0.4em, topsep=0.4em]
\item \textbf{Boundary marginalization:} a stable log-domain algorithm evaluates and differentiates interval likelihoods using at most two paths of length $d=\log_b M$ (Proposition~\ref{prop:exact}). 
\item \textbf{Learning at the observed resolution:} a single likelihood handles precise labels, intervals, and bounds, with a local gradient that characterizes the digit decisions supervised by each observation (Proposition~\ref{prop:resolution}). 
\item \textbf{Exact queries and proper scores:} the same computation supports exact tail and quantile queries and unbiased threshold-sampled CRPS gradients (Section~\ref{sec:crps}). 
\end{enumerate}
Experiments establish the value of incomplete evidence in controlled recovery, naturally censored molecular measurements, and code latency. They demonstrate learning without precise labels, reproduce the latency gains on a second backbone, and measure precision scalability.

\section{Related Work}
\paragraph{Numerical decoders and training through the numeric distribution.}
LLMTime treats digit sequences as a hierarchical numerical density and samples forecasts \citep{gruver2023}. Decoding-based regression studies numerical serialization, cross-entropy training, and decoder-based density estimation \citep{song2025}, and LLM Processes condition numerical prediction on natural-language context \citep{requeima2024}. Regression-aware inference estimates Bayes-optimal point predictions from sampled answers \citep{lukasik2024}, and regression-aware fine-tuning (RAFT) differentiates a squared loss through the expectation of the numeric token distribution over an enumerated grid of candidate values \citep{lukasik2025}. Reinforcement learning has been used to optimize sequence-level numeric rewards over sampled outputs \citep{chen2026} and CRPS-based distributional rewards over rollouts \citep{park2026}. Piskorz et al.\ recover distributional summaries from hidden states with probes trained on sampled outputs \citep{piskorz2026}. Arithmetic Sampling uses autoregressive prefix masses for arithmetic-code decoding \citep{vilnis2023}. Our main contribution is exact likelihood learning from incomplete numerical observations through boundary marginalization. CRPS uses the same computation to differentiate queried CDFs directly, without generating model outputs.

\paragraph{Digit-level numerical losses.}
Number Token Loss adds a Wasserstein or squared-error term over the number-token distribution at each position \citep{zausinger2025}. Numerical Token Integrity Loss (NTIL) combines token-level distance with a differentiable numerical point construction \citep{fei2025}. Digit Entropy Loss (DEL) reformulates digit-level entropy optimization with binary cross-entropy on digit conditionals \citep{zheng2026}, and Smooth Maximum Mean Discrepancy (SMMD) matches predicted and target numeric-token distributions using kernels based on distances between numerical values \citep{zuo2026}. Strictly proper non-logarithmic scoring rules have been applied to token conditionals \citep{shao2024}, and Calibration Fine-Tuning derives prefix targets from a specified output distribution \citep{baldelli2026}. This literature improves numerical token supervision and point constructions. Our objective instead assigns probability to the complete observed numerical event, including intervals and bounds for which no exact target numeral is supplied.

\paragraph{Coarsened labels.}
Heitjan and Rubin formalize observed-data likelihoods for coarsened data and the ignorability conditions under which they are valid \citep{heitjan1991}. Candidate-set likelihoods of the same form appear in learning with multiple labels \citep{jin2002}. Earlier numerical language models integrate a separate continuous mixture density over a numeral's rounding interval \citep{spithourakis2018}. Censored likelihood heads have been studied for uncertainty estimation in molecular property prediction \citep{svensson2025}. Recent language-model approaches to censored survival outcomes learn relative survival orderings from comparisons between subjects \citep{wei2026,xiang2026}. We train the model's numerical distribution directly on the censored likelihood through its existing digit tokens and compare against separate distributional output heads trained on the same likelihood (event-aware heads) as primary baselines.

\paragraph{Tree-structured numerical distributions.}
Hierarchical softmax variants over ordered bins compute leaf probabilities along a path and smooth neighboring leaves \citep{tansey2017}. PATH, a concurrent tabular method, learns hierarchical output distributions for interval prediction by combining a base leaf distribution with autoregressive branch refinement \citep{cai2026}. We instead fine-tune a pretrained numeral decoder on observed intervals and bounds, computing their event likelihood by boundary marginalization without constructing the leaf distribution. NumericJev decodes numbers from a structured choice interface at inference time without training \citep{ye2026}.

\section{Learning from Numerical Observations}
\label{sec:eventlearning}
We represent each observation by the grid values compatible with it and train the LM to assign probability to that set. A fixed-width digit representation places these values in numerical order.
\subsection{An ordered, fixed-width numerical distribution}
We represent numerical outputs on an ordered grid and encode each grid index with a fixed number of digits, starting with the most significant digit. Comparing these strings from left to right preserves numerical order. Let the output support be
\begin{equation}
\cV=\{v_0<\cdots<v_{M-1}\},\qquad M=b^d,
\label{eq:grid}
\end{equation}
where $b$ is the number of allowed digits, $d$ is the fixed output length, and $M$ is the number of possible grid values. We write $K$ for the grid index and $Y=v_K$ for its numerical value. The uniform grid $v_k=a+\Delta k$ with $\Delta>0$ is the default. The same computation supports any strictly increasing grid, including a log-spaced grid for positive targets spanning several orders of magnitude. Let $z(k)=(z_1,\ldots,z_d)$ be the base-$b$ expansion of $k$, padded with leading zeros, with $z_1$ the most significant digit.

We choose $b$ existing token identifiers (IDs) representing the digits. For input $x$ and digit prefix $s$ (the digits already supplied), let $\ell_\theta(x,s)_a$ be the corresponding LM logit. The next-digit distribution is a softmax restricted to the $b$ digit tokens,
\begin{equation}
 p_\theta(a\mid x,s)
 =\frac{\exp\ell_\theta(x,s)_a}
 {\sum_{c=0}^{b-1}\exp\ell_\theta(x,s)_c},
 \qquad a\in\{0,\ldots,b-1\},
\label{eq:masked}
\end{equation}
and the probability of a complete numeral is the product of its digit probabilities,
\begin{equation}
 p_\theta(k\mid x)
 =\prod_{j=1}^{d}p_\theta(z_j(k)\mid x,z_{<j}(k)).
\label{eq:joint}
\end{equation}
Normalization over the $b$ digit tokens and deterministic termination after $d$ digits ensure that \eqref{eq:joint} sums to one over the $M$ grid values. We call it the LM's constrained numeral distribution. The task prompt declares the mapping from grid indices to numerical values, the support, and the units. Numerals are built directly from token IDs, independently of whether the tokenizer would merge adjacent characters. The LM predicts the target numeral directly using its existing backbone and vocabulary.

\subsection{Observations as compatible sets}
A training observation is $(x,A)$, where $A\subseteq\{0,\ldots,M-1\}$ contains the grid indices compatible with what was observed. We call $A$ the observed \emph{event} and focus on contiguous intervals $A=[l,u]\cap\mathbb Z$. Table~\ref{tab:events} defines the common cases on a uniform grid. An exact value is represented at the declared modeling resolution. When the reporting precision is coarser than that resolution, the observation is represented by its rounding interval.

\begin{table}[ht]
\centering\small
\setlength{\tabcolsep}{5pt}
\begin{tabular*}{\linewidth}{@{\extracolsep{\fill}}lcll@{}}
\toprule
Observation            & Numerical event      & \multicolumn{2}{c@{}}{Compatible grid indices} \\
\midrule
Exact grid label $v_r$ & $Y=v_r$              & \multicolumn{2}{c@{}}{$l=u=r$} \\
Closed interval        & $L\leq Y\leq U$      & $l=\lceil(L-a)/\Delta\rceil$     & $u=\lfloor(U-a)/\Delta\rfloor$ \\
Rounded to width $h$   & $r-h/2\leq Y<r+h/2$  & $l=\lceil(r-h/2-a)/\Delta\rceil$ & $u=\lceil(r+h/2-a)/\Delta\rceil-1$ \\
Strict lower bound     & $Y>c$                & $l=\lfloor(c-a)/\Delta\rfloor+1$ & $u=M-1$ \\
Strict upper bound     & $Y<c$                & $l=0$                            & $u=\lceil(c-a)/\Delta\rceil-1$ \\
\bottomrule
\end{tabular*}
\caption{Event construction. Indices are intersected with the declared support. Rounding uses the half-open convention shown, and other tie rules are handled the same way. Endpoints are converted with decimal or rational arithmetic (Appendix~\ref{app:impl}).}
\label{tab:events}
\end{table}

\subsection{One likelihood for all observations}
An event's probability is the sum of the probabilities of its compatible grid values. We call this sum the \emph{event likelihood} and, for $A=[l,u]\cap\mathbb Z$, write $\logmass_\theta(l,u\mid x)$ for its logarithm. We train by minimizing the negative log event likelihood:
\begin{equation}
 \loss_{\mathrm{event}}(\theta;x,A)
 =-\log\sum_{k\in A}p_\theta(k\mid x)
 =-\logmass_\theta(l,u\mid x).
\label{eq:eventloss}
\end{equation}
A singleton reduces exactly to sequence negative log likelihood (NLL), rounded measurements supervise bins, and bounds supervise tails. All observation types can appear in the same minibatch with one loss. Section~\ref{sec:boundary} computes this objective and its gradient without enumerating the compatible numerals.

The event likelihood and cross-entropy on an imputed numeral have different targets. Suppose a reporting rule maps an observation $A$ to a representative $r(A)$. Cross-entropy on $r(A)$ learns the distribution of the representatives, whereas \eqref{eq:eventloss} learns the probabilities of the compatible events. For fixed rounding bins, both objectives can match the bin masses, but imputation additionally concentrates targets at the reported centers. Event likelihood imposes only the observed probability constraints. Mixed reporting resolutions and heterogeneous bounds make this distinction central to learning a numerical distribution.

\paragraph{Likelihood interpretation.}
Conditionally ignorable reporting means that, for fixed $x$, all values compatible with a report have the same probability of producing it \citep{heitjan1991}. The probability of report $A$ given the underlying index $k$ then has the form
\begin{equation}
 q(A\mid k,x)=c(A,x)\ind\{k\in A\},\qquad
 \sum_{A\ni k}c(A,x)=1,
\label{eq:coarsening}
\end{equation}
where $c$ is independent of $\theta$. The probability of observing report $A$ is then $c(A,x)\sum_{k\in A}p_\theta(k\mid x)$, whose negative log likelihood equals \eqref{eq:eventloss} up to the constant $-\log c(A,x)$. This includes deterministic rounding and conditionally independent censoring. Known reporting conditions can enter $x$. Appendix~\ref{app:coarse} characterizes the distributional information identified by these reports.

\section{Exact Boundary Marginalization}
\label{sec:boundary}
We evaluate \eqref{eq:eventloss} by summing complete prefix subtrees, using only the event's boundary paths.

\subsection{One path evaluates a tail probability}
For a threshold index $k$ with digit string $z=z(k)$, define the prefix mass, the probability that a numeral begins with the first $j-1$ threshold digits,
\begin{equation}
 P_\theta(z_{<j})=\prod_{i<j}p_\theta(z_i\mid x,z_{<i}),
 \qquad P_\theta(\varnothing)=1,
\end{equation}
where the input $x$ is omitted from symbols whose dependence on it is clear from context. The inclusive CDF is $F_\theta(k\mid x)=P_\theta(Y\leq v_k\mid x)$. Every numeral below the threshold matches $z$ up to some position $j$, takes a smaller digit there, and then continues with an unrestricted suffix whose total probability is one. Grouping numerals by that position and adding the threshold numeral itself gives
\begin{equation}
 F_\theta(k\mid x)
 =\sum_{j=1}^{d}P_\theta(z_{<j})
       \sum_{a<z_j}p_\theta(a\mid x,z_{<j})
       +p_\theta(k\mid x),
\label{eq:cdf}
\end{equation}
and the same grouping with larger digits gives the strict survival probability, the probability of exceeding the threshold,
\begin{equation}
 S_\theta(k\mid x)=P_\theta(Y>v_k\mid x)
 =\sum_{j=1}^{d}P_\theta(z_{<j})
       \sum_{a>z_j}p_\theta(a\mid x,z_{<j}).
\label{eq:survival}
\end{equation}
All $d$ conditional digit distributions in these expressions come from a single teacher-forced path, which a causal transformer evaluates in one forward pass. The threshold is given by the observation or query, so no model output is sampled.

\subsection{Stable interval mass from two paths}
Computing $F_\theta(u)-F_\theta(l-1)$ erases a small interval probability whenever both CDFs round to the same floating-point number. We therefore write the interval as a disjoint union of prefix subtrees and sum their masses in the log domain.

For $l<u$, let $l'=z(l)$ and $u'=z(u)$ be the boundary digit strings, and let $r$ be their first differing position, indexed from $1$. As illustrated in Figure~\ref{fig:tree}, the event consists of four disjoint groups: (1) the two boundary numerals, (2) subtrees that share the boundary prefix before $r$ and choose a digit strictly between the two boundary digits at $r$, (3) subtrees that first depart from the lower boundary by choosing a larger digit at a position after $r$, and (4) subtrees that first depart from the upper boundary by choosing a smaller digit at a position after $r$. Each included subtree admits every suffix, so its total probability is the probability of reaching its prefix. For a digit string $t$, define its log prefix masses and the log probability of the complete string,
\begin{equation}
 H_j(t)=\sum_{i<j}\log p_\theta(t_i\mid x,t_{<i}),
 \qquad H_{d+1}(t)=\log p_\theta(t\mid x).
\end{equation}
For sibling subtrees at a shared prefix, the total probability is the prefix mass times the sum of the included next-digit probabilities. Collecting the log masses of the boundary numerals and these disjoint subtree groups in the same order gives
\begin{align}
\mathcal T(l,u)= {}& \bigl(H_{d+1}(l'),H_{d+1}(u')\bigr)\nonumber\\
&\mathbin{\Vert}\left(H_r(l')+\log\sum_{a=l'_r+1}^{u'_r-1}
                 p_\theta(a\mid x,l'_{<r})\right)\nonumber\\
&\mathbin{\Vert}
 \left(H_j(l')+\log\sum_{a>l'_j}p_\theta(a\mid x,l'_{<j})\right)_{j=r+1}^{d}\nonumber\\
&\mathbin{\Vert}
 \left(H_j(u')+\log\sum_{a<u'_j}p_\theta(a\mid x,u'_{<j})\right)_{j=r+1}^{d},
\label{eq:cover}
\end{align}
where $\Vert$ denotes tuple concatenation and empty digit ranges are omitted. With $\LSE$ denoting log-sum-exp, the interval log probability is
\begin{equation}
 \logmass_\theta(l,u\mid x)=\LSE\bigl(\mathcal T(l,u)\bigr).
\label{eq:logmass}
\end{equation}
For $l=u$, the log probability is $H_{d+1}(z(l))$. For the full support it is zero. One-sided intervals use $\logmass_\theta(0,u)=\log F_\theta(u)$ or $\logmass_\theta(l,M-1)=\log S_\theta(l-1)$ for $l>0$, so one path suffices. An empty event is rejected. For $237\leq Y\leq684$ in Figure~\ref{fig:tree}, these four groups are $\{237,684\}$, $\{3**,4**,5**\}$, $\{24*,\ldots,29*,238,239\}$, and $\{60*,\ldots,67*,680,\ldots,683\}$. Every required next-digit distribution is evaluated at a prefix of $237$ or $684$.

\begin{proposition}[Exactness and query complexity]
For \eqref{eq:joint}, equations \eqref{eq:cdf}--\eqref{eq:logmass} compute the exact event probabilities in real arithmetic. One tail uses at most $d$ prefix conditionals and one interval at most $2d$. Grouped interval evaluation uses at most $2d+1$ log-mass terms and $O(bd)=O(b\log_b M)$ scalar arithmetic, with no enumeration of the support or interval interior. Automatic differentiation gives the exact gradients of the event probability and event log likelihood.
\label{prop:exact}
\end{proposition}

\subsection{Which digits an interval supervises}
On the three-digit grid $v_k=k$, the event $440\leq Y\leq449$ consists of every completion of \texttt{44}. Its probability is simply the prefix probability: the loss supervises the first two digits but supplies no target for the third.

More generally, let $p_\theta(k\mid x,A)$ be the model distribution restricted to compatible indices and renormalized to sum to one. The gradient of the event loss is the negative average gradient of the compatible numerals' log probabilities, weighted by their probabilities conditional on $A$:
\begin{equation}
 \nabla_\theta\loss_{\mathrm{event}}
 =-\E_{K\sim p_\theta(\cdot\mid x,A)}
            [\nabla_\theta\log p_\theta(K\mid x)].
\label{eq:posteriorgradient}
\end{equation}
Boundary marginalization computes this average exactly along the boundary paths without enumerating or sampling compatible numerals.

For a prefix $s$, let $C_s$ contain the indices whose numerals extend $s$, and let $sa$ denote that prefix followed by digit $a$. Write $\ell_s(a)=\ell_\theta(x,s)_a$ for its next-digit logit. Conditioned on the event $A$, let $w_s$ be the posterior probability of reaching prefix $s$ and $t_s(a)$ the posterior probability that the next digit after $s$ is $a$,
\begin{equation}
 w_s=P_\theta(C_s\mid A,x),\qquad
 t_s(a)=P_\theta(C_{sa}\mid C_s,A,x),
\end{equation}
with the convention $t_s=p_\theta(\cdot\mid x,s)$ when $w_s=0$. The derivative with respect to the next-digit logits at $s$ then has the usual softmax cross-entropy form, with $t_s$ as the target distribution and $w_s$ as its weight.

\begin{proposition}[Gradient at the observed resolution]
Holding the logits at other prefixes fixed, the derivative with respect to the next-digit logit $\ell_s(a)$ is
\begin{equation}
 \frac{\partial\loss_{\mathrm{event}}}{\partial\ell_s(a)}
 =w_s\bigl[p_\theta(a\mid x,s)-t_s(a)\bigr].
\label{eq:localgradient}
\end{equation}
If $C_s\subseteq A$ or $C_s\cap A=\varnothing$, this derivative is zero. Consequently, an interval event directly supervises only prefix nodes whose subtrees straddle a boundary, and $|\partial\loss_{\mathrm{event}}/\partial\ell_s(a)|\leq w_s\leq1$. The chain rule propagates these derivatives to the shared LM parameters $\theta$.
\label{prop:resolution}
\end{proposition}

The two boundary paths contain every prefix that can receive a nonzero local gradient. Boundary marginalization therefore preserves the exact gradient without visiting the interval interior.

\paragraph{What the labels identify.}
Event likelihood matches the distribution of the observed reports. Reports from fixed, disjoint rounding bins identify the total probability of each bin, and finer or overlapping reports can additionally distinguish probabilities within a bin. Appendix~\ref{app:coarse} states this as an identity for the expected loss and gives the identifiability condition. This motivates our controlled experiment, which varies the precise-label probability down to zero while retaining rounding intervals and bounds $Y\leq c$ or $Y>c$ at thresholds sampled independently of $Y$.

\section{Implementation, Inference, and CRPS Training}
The same boundary computation supports event likelihoods, numerical queries, and CRPS training. We describe how teacher forcing supplies the required digit probabilities and how each task uses them.
\subsection{Teacher-forced probability extraction}
Given prompt tokens $x_1,\ldots,x_n$ and threshold digits $z_1,\ldots,z_d$, feed the prompt followed by $z_1,\ldots,z_{d-1}$. Read causal next-token logits at zero-based positions $n-1,\ldots,n+d-2$, select the $b$ digit-token columns, and apply log-softmax in 32-bit floating-point (FP32) arithmetic. This provides all $d$ conditional distributions for one path. The selected paths are evaluated together in a single batched forward pass. Appendix~\ref{app:verify} verifies this computation and its gradients against exhaustive enumeration.

Singleton and one-sided events require one path, interior intervals at most two, and full-support events contribute zero loss. Dropout is disabled so shared prefixes define one coherent distribution, while automatic differentiation remains enabled.

\paragraph{Cost.}
Boundary marginalization combines one or two paths of depth $d=\log_b M$ using $O(bd)$ scalar arithmetic. Our reference implementation repeats the prompt for each path and uses ordinary causal attention. A layout that shares the prompt and the common digit prefix between the two paths adds only $d-r\leq d-1$ input positions for the second boundary, where $r$ is the first differing digit. The method supports full fine-tuning and parameter-efficient updates such as low-rank adaptation (LoRA).

\subsection{Algorithmic summary}
Algorithm~\ref{alg:event} summarizes one training update from observed events.

\begin{algorithm}[ht]
\caption{Event-supervised numerical update}
\label{alg:event}
\begin{enumerate}[leftmargin=*, topsep=2pt, itemsep=2pt]
\item Convert each observation into a nonempty compatible interval $[l,u]$ on the declared grid.
\item Return zero loss for the full support. Otherwise select the singleton path, the tail threshold, or the two interval boundaries.
\item Teacher-force the selected paths with repeated prompts and extract locally normalized digit log probabilities.
\item Use the appropriate log tail or sum the subtree masses in \eqref{eq:cover} with log-sum-exp.
\item Average $-\logmass_\theta(l,u\mid x)$ over examples and backpropagate through the selected paths.
\end{enumerate}
\smallskip
{\small Event-likelihood training by boundary marginalization. The CRPS update uses the same computation (Algorithm~\ref{alg:crps}, Appendix~\ref{app:crpsimpl}).}
\end{algorithm}

\subsection{Exact numerical queries at inference}
The trained model answers distributional queries without sampling. A tail probability $P(Y>c\mid x)$, for example the probability that a program exceeds a latency budget, is one boundary path in one forward pass. An interval probability uses at most two paths. The $\alpha$-quantile is the smallest grid value with cumulative probability at least $\alpha$. We find it by choosing digits from most to least significant. Initialize $\tau=\alpha$ for the desired level $\alpha\in(0,1)$. At each prefix, choose the first digit $a$ whose cumulative child probability reaches $\tau$. To express the target level within the chosen subtree, subtract the mass of preceding sibling subtrees and divide by the chosen digit's probability:
\begin{equation}
 \tau\leftarrow
 \frac{\tau-\sum_{c<a}p_\theta(c\mid x,s)}{p_\theta(a\mid x,s)}.
\end{equation}
After $d$ sequential digit steps, this traversal returns the lower $\alpha$-quantile, with a fixed convention at probability boundaries. The same quantile can be obtained by deterministic CDF bisection with at most $\lceil\log_2 M\rceil$ boundary queries, as used for the reported medians. An equal-tailed prediction interval uses two quantile searches.

\subsection{Threshold-sampled CRPS training}
\label{sec:crps}
Boundary marginalization also supports CRPS training on exact labels. CRPS integrates the squared difference between the predicted CDF and the step function that jumps from zero to one at the observed value.
For an exact grid label $y=v_r$, define $h_k(r)=\ind\{r\leq k\}$ and $\Delta_k=v_{k+1}-v_k$. The CRPS of the discrete output distribution is
\begin{equation}
 \loss_{\mathrm{CRPS}}(\theta;x,v_r)
 =\sum_{k=0}^{M-2}\Delta_k
       \bigl[F_\theta(k\mid x)-h_k(r)\bigr]^2.
\label{eq:crps}
\end{equation}

We estimate this sum by sampling a few thresholds and weighting each term by the inverse of its sampling probability. Choose a sampling distribution $\pi(k\mid x,y)>0$ over thresholds $k\in\{0,\ldots,M-2\}$ that is fixed with respect to $\theta$. For independent thresholds $K_1,\ldots,K_m\sim\pi$, use
\begin{equation}
 \widehat\loss_{\mathrm{CRPS}}
 =\frac1m\sum_{j=1}^{m}
 \frac{\Delta_{K_j}}{\pi(K_j\mid x,y)}
 \bigl[F_\theta(K_j\mid x)-h_{K_j}(r)\bigr]^2.
\label{eq:samplecrps}
\end{equation}
Each sampled threshold requires one teacher-forced boundary path. Both the loss estimator and its ordinary backpropagation gradient are unbiased for the full CRPS (Proposition~\ref{prop:crps}). Appendix~\ref{app:crpsimpl} gives the numerical-distance interpretation, the sampling choices, stable residual evaluation, and Algorithm~\ref{alg:crps}. Proposition~\ref{prop:rb} establishes the variance reduction relative to a matched estimator that samples model outputs.

\section{Experiments}
\label{sec:experiments}
We test how event likelihood converts incomplete observations into distributional supervision. Three settings cover recovery of a known conditional distribution, naturally censored molecular reports, and code latency on the subset of Code-Regression labeled KBSS, which contains Triton kernels for GPUs \citep{coderegression}. The latency study uses constructed training timeouts and fully observed test outcomes. We then test a second backbone, compare proper scores on exact labels, and measure precision scalability.

The primary backbone is Qwen2.5-1.5B, and the cross-backbone study uses Qwen3-4B-Base. All learning comparisons use rank-16 LoRA, a learning rate of $10^{-4}$, and a four-digit numerical grid.

We compare five methods on incomplete labels (Table~\ref{tab:baselines}, Appendix~\ref{app:exp}). Event likelihood is the proposed objective. Complete-case CE trains the same decoder with sequence CE on the precise labels only, and imputed-label CE trains it on one representative numeral per report. The two event-aware heads keep the same backbone and the same event likelihood but replace the digit decoder with a five-component Gaussian mixture or a 256-bin histogram output. The CE controls test the value of incomplete reports and their event representation, and the heads compare output parameterizations given the same reports.

Within each comparison, methods share backbone initialization, adapter placement, optimizer settings, update budget, and validation schedule. The head baselines add their declared output parameters, and complete-case CE repeats its precise subset to match the update budget. The incomplete-label studies select checkpoints by event NLL across all validation reports. The exact-label score comparison selects by validation CRPS. Appendix~\ref{app:completedresults} provides supporting scores, individual runs, and integration checks.

\subsection{Learning from incomplete numerical evidence}
\paragraph{Controlled distribution recovery.}
We construct 8,000/1,000/2,000 training/validation/test examples with inputs drawn uniformly on $[-1,1]$ and numerical outcomes drawn from an input-dependent two-component Gaussian mixture truncated to $[-2,2]$. The same latent draws are reported with nominal precise-label probabilities $\rho\in\{0,0.1,0.2,0.5,1\}$. The remaining observations provide rounding intervals or one-sided bounds at thresholds sampled independently of the outcome. Integrated squared CDF error (ISE) measures recovery by integrating the squared difference between the predicted and true conditional CDFs over the support.

\begin{figure}[t]
\centering
\paperfigure{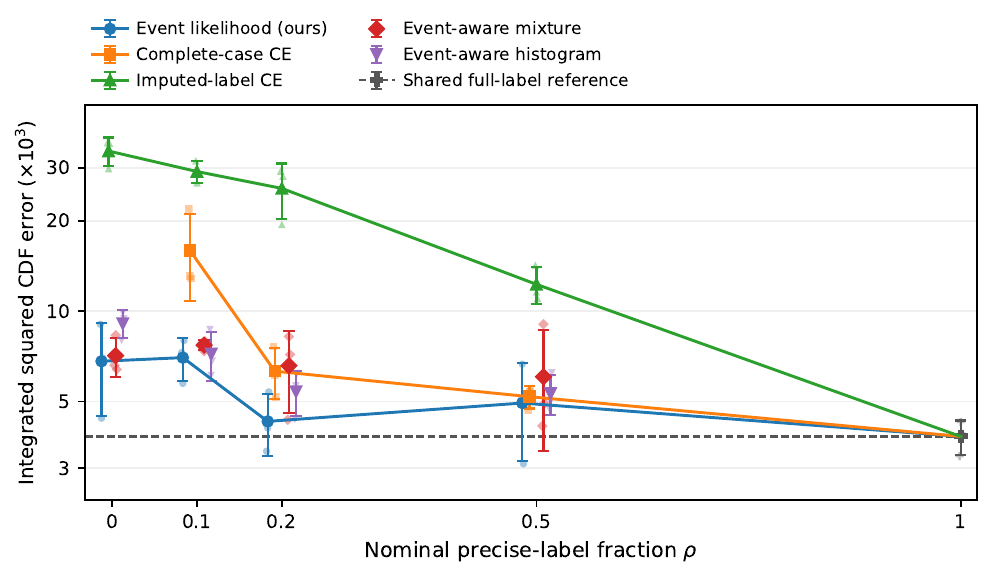}{.90\linewidth}
\caption{Distribution recovery across precise-label probabilities. Large markers show mean ISE and error bars show one sample standard deviation (SD) over seeds 0, 1, and 2. Faint markers show individual seeds, offset horizontally for readability. Lower is better. The horizontal axis is linear and the vertical axis logarithmic. Event-aware heads are evaluated at $\rho\in\{0,0.1,0.2,0.5\}$, while complete-case CE is unavailable at $\rho=0$. At $\rho=1$, the three decoder objectives share one full-label reference, whose mean is marked by the dashed line. ISE uses 64 shared stratified thresholds per test input (Appendix~\ref{app:exp}). Results for individual seeds appear in Figure~\ref{fig:recoveryseeds}.}
\label{fig:recovery}
\end{figure}

Figure~\ref{fig:recovery} compares recovery across precise-label probabilities, with ISE values reported below in units of $10^{-3}$. As precise labels become scarce, complete-case CE degrades while event likelihood remains within a factor of two of the full-label reference, $3.819$ at $\rho=1$. At $\rho=0.1$, event likelihood achieves mean ISE $6.993$, compared with $15.923$ for complete-case CE and $29.156$ for imputed-label CE, reducing error by 56.1\% relative to complete-case CE. It improves on both CE controls in all three paired seeds. With no precise training labels ($\rho=0$), rounding and threshold reports alone yield mean ISE $6.801$, an 80.1\% reduction from imputed-label CE ($34.121$). Complete-case CE is unavailable in this setting.

Event likelihood has lower mean ISE than both event-aware heads at every $\rho\in\{0,0.1,0.2,0.5\}$ while using the model's existing digit tokens. At $\rho=0.5$, its mean ISE is $4.936$, compared with $6.037$ for the mixture head and $5.312$ for the histogram head. Imputation performs worse than discarding incomplete reports at $\rho=0.1$, $0.2$, and $0.5$, consistent with Section~\ref{sec:eventlearning}: cross-entropy on imputed numerals learns the distribution of the representatives. The five-seed follow-up at $\rho=0.2$ gives mean ISE $4.818$ for event likelihood and $7.906$ for complete-case CE, a 39.1\% reduction.

\begin{figure}[t]
\centering
\paperfigure{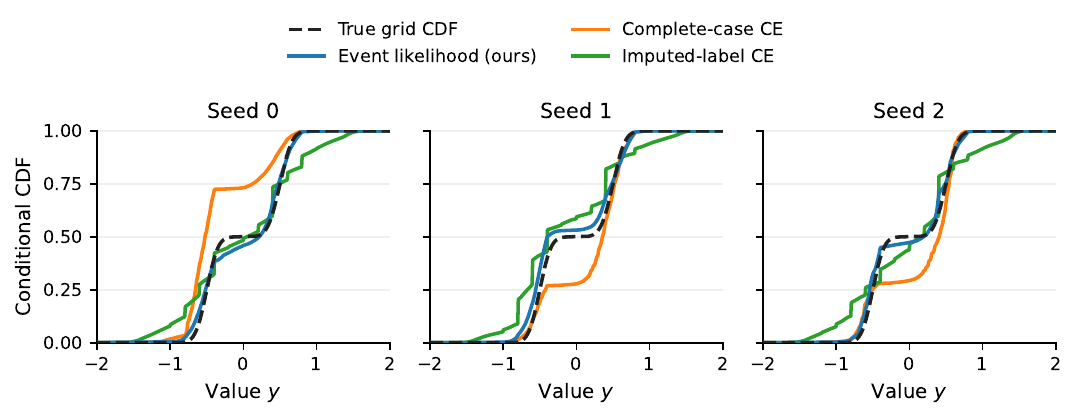}{\linewidth}
\caption{Conditional CDFs at $\rho=0.1$ for the test input nearest $x=0$ ($x=-0.000416$), with one panel per training seed. Curves use checkpoints selected by validation event NLL and connect evaluations at 512 common grid thresholds. The dashed curve is the true grid CDF.}
\label{fig:conditionalcdf}
\end{figure}

At the test input nearest $x=0$, event likelihood captures both rises and the central plateau of the true conditional CDF in all three seeds (Figure~\ref{fig:conditionalcdf}). Complete-case CE produces central plateaus near $0.73$, $0.28$, and $0.29$ for seeds 0, 1, and 2, compared with the true $0.50$. Imputed-label CE shows large jumps near rounding bin centers and broader tails, consistent with its use of representative numerals as targets (Section~\ref{sec:eventlearning}).

\paragraph{Naturally censored molecular measurements.}
We retain 14,850 IC50 measurements of human ether-\`a-go-go-related gene (hERG) channel inhibition, a standard cardiac safety assay, from ChEMBL 37 \citep{chembl37}. We transform concentrations and bounds to pIC50, the negative base-10 logarithm of molar IC50. We split the reports into 11,880 training, 1,485 validation, and 1,485 test reports so that no parent molecule or scaffold (core molecular structure) is shared across splits. Bounds constitute 29.8\% of training reports and 36.2\% of test reports, and the remaining reports are equalities, which give an exact value.

Event likelihood incorporates 3,535 bounded training reports excluded by complete-case CE and achieves the lowest NLL over all test reports, 4.675, improving over complete-case CE by 0.408 nats per report and over both event-aware heads (Table~\ref{tab:mainresults}). Its NLL on censored reports is 0.823 versus 2.076 for complete-case CE, a 60.3\% reduction. On equality reports, its NLL is below imputed-label CE and both heads, and 0.071 nats higher than complete-case CE (6.857 versus 6.786), which trains only on equality reports. Table~\ref{tab:hergsupport} (Appendix~\ref{app:completedresults}) reports NLL by bound type, equality CRPS, and the mean absolute error (MAE) of predicted medians.

\begin{table}[t]
\centering\small
\setlength{\tabcolsep}{4pt}
\begin{tabular*}{\linewidth}{@{\extracolsep{\fill}}lcccc@{}}
\toprule
& \multicolumn{2}{c}{hERG: natural reports} & \multicolumn{2}{c}{KBSS: exact test outcomes} \\
\cmidrule(lr){2-3}\cmidrule(l){4-5}
Method & Full NLL & Equality NLL & CRPS ($10^{-3}$ ms) & CRPS ($\log_{10}$ ms) \\
\midrule
Event likelihood (ours) & $\mathbf{4.675}\!\pm\!0.012$ & $6.857\!\pm\!0.049$ & $\mathbf{8.262}\!\pm\!0.299$ & $\mathbf{0.0878}\!\pm\!0.0013$ \\
Complete-case CE & $5.083\!\pm\!0.055$ & $\mathbf{6.786}\!\pm\!0.062$ & $8.870\!\pm\!0.235$ & $0.0980\!\pm\!0.0024$ \\
Imputed-label CE & $4.958\!\pm\!0.032$ & $6.998\!\pm\!0.086$ & $10.229\!\pm\!0.135$ & $0.1205\!\pm\!0.0026$ \\
Event-aware mixture & $5.338\!\pm\!0.023$ & $7.880\!\pm\!0.027$ & $11.374\!\pm\!0.626$ & $0.0973\!\pm\!0.0031$ \\
Event-aware histogram & $5.413\!\pm\!0.087$ & $7.934\!\pm\!0.053$ & $8.990\!\pm\!0.780$ & $0.0895\!\pm\!0.0022$ \\
\bottomrule
\end{tabular*}
\caption{Incomplete-label learning on real inputs with Qwen2.5-1.5B. Entries are means $\pm$ sample standard deviation (SD) over seeds 0, 1, and 2. Lower is better, and bold marks the lowest mean. hERG full NLL covers all 1,485 test reports (948 equalities and 537 bounds), and equality NLL covers the equalities only. KBSS uses all 744 exact test latencies. Methods share prompts, grid, update budget, and checkpoint selection by validation event NLL within each task.}
\label{tab:mainresults}
\end{table}

\paragraph{Code latency from timeout supervision.}
KBSS pairs each Triton GPU kernel with its measured latency. We split it by code template into 5,949 training, 744 validation, and 744 test examples. Independently assigned timeouts censor 45.2\% of training outcomes. Every test latency is retained exactly.

Event likelihood obtains the lowest CRPS on both scales: $8.262\times10^{-3}$ ms and 0.0878 in $\log_{10}(\mathrm{ms})$. Physical and logarithmic CRPS decrease by 6.8\% and 10.5\% relative to complete-case CE, and by 8.1\% and 1.9\% relative to the event-aware histogram. Both CRPS scores improve against every control in each of the three paired seeds (Figure~\ref{fig:pairedeffects}, Appendix~\ref{app:completedresults}). For the event that a kernel exceeds a fixed latency budget, the Brier score measures the mean squared error between the predicted probability and the observed binary outcome. It is 0.1281 versus 0.1550 for complete-case CE, showing more accurate predictions for this tail event (Table~\ref{tab:latencysupport}).

\subsection{Transfer to a newer, larger backbone}
We repeat the event, complete-case CE, and histogram comparison on Qwen3-4B-Base \citep{qwen3}, preserving the KBSS split, timeout assignments, prompts, numerical support, 558-update budget, and validation rule. Event likelihood again achieves the lowest CRPS on both scales (Table~\ref{tab:crossbackbone}). Physical and logarithmic CRPS decrease by 5.3\% and 11.5\% relative to complete-case CE, and by 4.3\% and 4.0\% relative to the histogram head. Its budget Brier score is also lowest. These results reproduce the benefit of event supervision on a second pretrained backbone.

\begin{table}[ht]
\centering\small
\begin{tabular}{@{}lccc@{}}
\toprule
Method & CRPS ($10^{-3}$ ms) & CRPS ($\log_{10}$ ms) & Budget Brier \\
\midrule
Event likelihood (ours) & $\mathbf{8.395}\!\pm\!0.274$ & $\mathbf{0.0878}\!\pm\!0.0018$ & $\mathbf{0.1327}\!\pm\!0.0059$ \\
Complete-case CE & $8.862\!\pm\!0.119$ & $0.0992\!\pm\!0.0031$ & $0.1541\!\pm\!0.0067$ \\
Event-aware histogram & $8.776\!\pm\!0.488$ & $0.0915\!\pm\!0.0008$ & $0.1409\!\pm\!0.0036$ \\
\bottomrule
\end{tabular}
\caption{KBSS transfer to Qwen3-4B-Base. Entries are means $\pm$ sample SD over seeds 0, 1, and 2 on all 744 exact test outcomes. Both CRPS scales and the evaluation budget fixed from the training data match the Qwen2.5 evaluation. Lower is better, and bold marks the lowest mean. Table~\ref{tab:crosssupport} reports median MAE on both scales.}
\label{tab:crossbackbone}
\end{table}

\subsection{Exact-label score comparison}
Boundary marginalization also makes CRPS trainable (Section~\ref{sec:crps}), so we test whether CRPS should replace the likelihood as the training objective when every label is exact. With exact labels, event likelihood reduces to sequence CE, which supervises all $d$ digit conditionals of the observed numeral without sampling thresholds. On the fully precise controlled task, we compare sequence CE with threshold-sampled CRPS and binary cross-entropy on threshold events (CDF-BCE) through the same decoder at a common 2,250-update budget, with checkpoints selected by validation CRPS for every method.

Sequence CE attains the lowest mean test CRPS (0.2165), ISE, and point NLL (the NLL of the complete precise test numeral), supporting event likelihood as the primary training objective (Table~\ref{tab:scores}, Appendix~\ref{app:completedresults}). Querying only $m=2$ of the $M-1=9{,}999$ grid thresholds per observation, threshold CRPS achieves mean test CRPS 0.2197, about 1.5\% above sequence CE. It has lower means than CDF-BCE (0.2219 test CRPS) on all three metrics. The same boundary computation thus supports direct CRPS training without generated rollouts.

\subsection{Precision scalability}
We hold physical support and interval observations fixed while increasing decimal depth from $d=2$ to $6$, expanding the output support from $10^2$ to $10^6$ values. Event update time and peak memory remain nearly constant across the tested depths. With 256 attended prompt tokens, updates take 0.872--0.888 seconds and use 11.224--11.343 GiB of peak allocated memory. With 1,024 tokens, they take 2.102--2.144 seconds and use 27.008--27.458 GiB. Exact interval training therefore remains tractable across this $10^4$-fold increase in support size.

\begin{figure}[ht]
\centering
\paperfigure{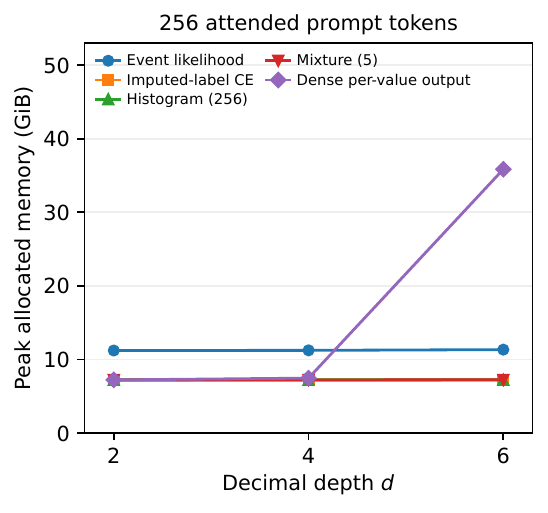}{.48\linewidth}\hfill
\paperfigure{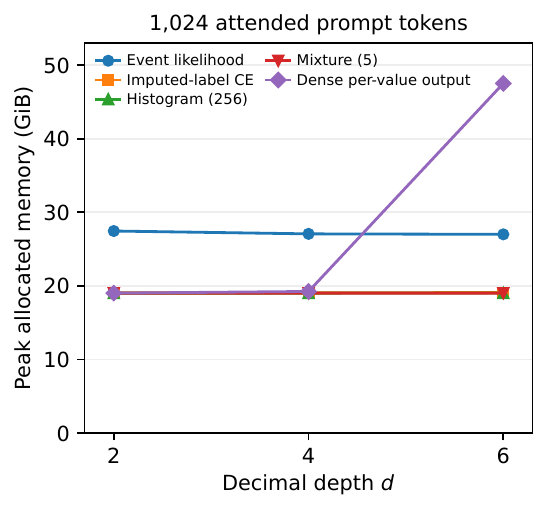}{.48\linewidth}
\caption{Precision scaling without dense output growth. Depths $2,4,6$ represent $10^2,10^4,10^6$ values on the same support. Methods share inputs and optimizer settings. Event likelihood and the three heads use the same interval reports, and imputed-label CE trains with sequence CE on one representative numeral per report. Memory is maximum peak allocated memory across three benchmark replicates, including weights, gradients, and Adam state. The dense per-value output predicts one probability per grid value, and the histogram and mixture heads have fixed output dimensions.}
\label{fig:precision}
\end{figure}

At $d=6$, the dense per-value output, which predicts one probability per grid value, uses 35.849 and 47.513 GiB for the two prompt lengths. Event likelihood uses 68.4\% and 43.2\% less memory, respectively (Figure~\ref{fig:precision}). Event likelihood updates take 1.07--1.10 times as long as imputed-label CE updates with 256 prompt tokens and 1.45--1.47 times as long with 1,024 tokens. Appendix~\ref{app:precisionresults} reports the full grid, including fixed-size event-aware heads.

\FloatBarrier
\section{Conclusion}
Boundary marginalization makes numerical events directly trainable in an autoregressive numeral distribution. Precise labels, rounding intervals, and censored outcomes contribute their exact likelihood through at most two boundary paths. Controlled experiments show gains over complete-case training when precise labels are scarce and over imputation when only rounding and threshold reports are available. Naturally censored molecular measurements and code latency further demonstrate the value of event supervision, with the latency gains reproduced on a second backbone. The same computation provides exact distributional queries and direct CRPS training, and the precision benchmark establishes tractable interval updates across a million-value support. Numerical supervision can therefore preserve the information supplied by each observation throughout learning and prediction.

\appendix

\clearpage
\setcounter{topnumber}{4}
\setcounter{bottomnumber}{3}
\setcounter{totalnumber}{6}
\renewcommand{\topfraction}{0.92}
\renewcommand{\bottomfraction}{0.85}
\renewcommand{\textfraction}{0.08}
\renewcommand{\floatpagefraction}{0.75}
\makeatletter
\setlength{\@fptop}{0pt}
\setlength{\@fpsep}{16pt plus 2pt minus 2pt}
\setlength{\@fpbot}{0pt plus 1fil}
\makeatother

\begin{center}
    \Large \textbf{Appendix}
\end{center}

\section{Limitations}
\paragraph{Finite representation and shared parameters.}
The declared grid fixes the support and resolution of the predictive distribution. Proposition~\ref{prop:resolution} characterizes derivatives with respect to individual prefix logits. Shared LM parameters can still couple predictions across prefix subtrees. Threshold-sampled scores retain sampling variance. The variance reduction in Proposition~\ref{prop:rb} compares with a matched estimator that samples model outputs.

\paragraph{Reporting and identifiability.}
The likelihood interpretation of \eqref{eq:eventloss} assumes conditionally ignorable reporting \eqref{eq:coarsening}. Nonignorable reporting requires modeling the reporting mechanism. Observed reports identify the event probabilities characterized by Proposition~\ref{prop:coarse}. At $\rho=0$, the controlled rounding and threshold design leaves some individual grid probabilities in the tails unidentified.

\paragraph{Empirical scope.}
Predictive rankings depend on the score: the mixture head has lower point NLL on the exact test values at $\rho=0$ (Table~\ref{tab:controlledall}). Outcomes behind natural hERG bounds are unobserved, and the assay reporting process selects the hERG equality subset. KBSS uses constructed timeouts and code templates as proxies for related programs. The two Qwen backbones differ in both generation and size, and seed variation on fixed splits does not quantify uncertainty from sampling new data. Predictive accuracy across grid depths remains untested.

\section{Statistical Properties and Proofs}
\subsection{Proof of Proposition~\ref{prop:exact}}
For a threshold string $z$, partition the complete strings smaller than $z$ by their first differing position $j$. Such a string matches $z_{<j}$, takes a digit $a<z_j$, and has an arbitrary suffix. Its partition mass is
\[
 P_\theta(z_{<j})p_\theta(a\mid x,z_{<j})
 \sum_{t\in\{0,\ldots,b-1\}^{d-j}}p_\theta(t\mid x,z_{<j},a).
\]
The suffix sum is one by repeated normalization of \eqref{eq:masked}. These partitions are disjoint and exhaustive. Adding the threshold leaf proves \eqref{eq:cdf}, and using $a>z_j$ proves \eqref{eq:survival}.

For two different boundaries, all compatible strings share the boundary prefix at positions before $r$. A digit strictly between $l'_r$ and $u'_r$ admits every suffix. A digit equal to $l'_r$ must remain at least the lower suffix: it either equals the lower leaf or first becomes larger at a later position. The upper branch is symmetric. This is exactly \eqref{eq:cover}, with no overlap. The largest collection has two leaves, one middle group, and at most $2(d-r)$ later groups, hence at most $2d+1$ terms.

Only conditional distributions on the two boundary strings occur, so there are at most $2d$ queried prefix nodes. Each grouped sibling sum examines at most $b$ digits, which establishes $O(bd)$ arithmetic. The sums and products are finite differentiable expressions for the same function of $\theta$, so differentiating the identity proves gradient equality. Log-sum-exp changes the numerical representation and leaves the mathematical function unchanged.

\subsection{Proof of Proposition~\ref{prop:resolution}}
Holding the logits at other prefixes fixed, the log probability of a complete numeral $k$ has derivative
\[
 \frac{\partial\log p_\theta(k)}{\partial\ell_s(a)}
 =\ind\{k\in C_s\}\bigl[\ind\{k\in C_{sa}\}-p_\theta(a\mid x,s)\bigr].
\]
Differentiating the log sum over $A$ gives \eqref{eq:posteriorgradient}. Taking the negative posterior expectation of the displayed derivative yields
\[
 -P(C_{sa}\mid A)+P(C_s\mid A)p_\theta(a\mid x,s)
 =w_s[p_\theta(a\mid x,s)-t_s(a)].
\]
If $C_s\subseteq A$, conditioning on $A$ imposes no further restriction within $C_s$, so $t_s(a)=p_\theta(a\mid x,s)$. If $C_s\cap A=\varnothing$, $w_s=0$. For an ordered interval, a subtree can be partially inside the event only if it contains a boundary. This identifies which prefixes can receive nonzero local gradients. The chain rule propagates these derivatives to shared model parameters.

\subsection{Observed-distribution consistency}
\label{app:coarse}
For fixed $x$, let $B_{A,k}=c(A,x)\ind\{k\in A\}$ be the matrix that maps a distribution over values to the distribution over reported events, and let $p^*$ be the true grid distribution, so $Bp^*$ is the observable event distribution. Let $\mathcal R(p)$ be the expected event loss and $\KL$ denote Kullback--Leibler divergence.

\begin{proposition}[Observed-distribution consistency]
The excess expected event loss equals the KL divergence between the true and predicted report distributions:
\begin{equation}
 \mathcal R(p)-\mathcal R(p^*)=\KL(Bp^*\,\|\,Bp).
\label{eq:observedkl}
\end{equation}
Population minimizers therefore match the observable event distribution. The report probabilities uniquely determine the grid distribution exactly when every change in the grid distribution changes at least one report probability:
\begin{equation}
 \ker B\cap\Bigl\{\delta:\sum_{k=0}^{M-1}\delta_k=0\Bigr\}=\{0\}.
\label{eq:identify}
\end{equation}
\label{prop:coarse}
\end{proposition}

\begin{proof}
The expected loss is
\[
 \mathcal R(p)=-\sum_A(Bp^*)_A\log\sum_{k\in A}p_k.
\]
Since $(Bp)_A=c(A,x)\sum_{k\in A}p_k$, the difference from the corresponding risk at $p^*$ cancels every nonzero reporting factor and becomes
\[
 \sum_A(Bp^*)_A\log\frac{(Bp^*)_A}{(Bp)_A}
 =\KL(Bp^*\|Bp).
\]
Terms with zero true probability contribute zero. Thus minimizers match $Bp^*$. Two simplex distributions have the same observable event distribution exactly when their difference is in $\ker B$ and has zero sum. Condition~\eqref{eq:identify} makes this difference zero. If that condition fails, a nonzero null vector whose entries sum to zero perturbs an interior simplex point in both directions without changing the observable event distribution, so the map is not injective on the whole simplex.

\end{proof}

\subsection{CRPS propriety and unbiased differentiation}
\label{app:crps}
\begin{proposition}[Proper objective and unbiased optimization]
Let $F^*$ be the CDF of the conditional data distribution. The excess expected CRPS is the integrated squared difference between the predicted and true CDFs:
\begin{equation}
 \E_{Y\sim p^*}\loss_{\mathrm{CRPS}}(p,Y)
 -\E_{Y\sim p^*}\loss_{\mathrm{CRPS}}(p^*,Y)
 =\sum_{k=0}^{M-2}\Delta_k(F(k)-F^*(k))^2,
\label{eq:proper}
\end{equation}
so the score is strictly proper on the finite ordered support. Moreover, \eqref{eq:samplecrps} is unbiased for \eqref{eq:crps}, and its ordinary backpropagation gradient is unbiased for the gradient of \eqref{eq:crps}.
\label{prop:crps}
\end{proposition}

\begin{proof}
At threshold $k$, the random indicator $H_k=\ind\{Y\leq v_k\}$ has expectation $F^*(k)$. Expanding a squared loss gives
\[
 \E(F(k)-H_k)^2=(F(k)-F^*(k))^2+F^*(k)(1-F^*(k)).
\]
The second term does not depend on $F$. Weighted summation proves \eqref{eq:proper}. All weights are positive and all $M-1$ cumulative probabilities determine the $M$ grid masses, which establishes strict propriety.

For the sampled objective, each threshold's proposal probability cancels its inverse-probability weight:
\[
 \E_K\frac{\Delta_K}{\pi(K)}(F_\theta(K)-h_K)^2
 =\sum_k\Delta_k(F_\theta(k)-h_k)^2.
\]
The support is finite and $\pi$ is independent of $\theta$, so differentiation commutes with this finite expectation. Averaging independent draws preserves the identities. The same cancellation applies within each stratum in \eqref{eq:strata}.

\end{proof}

\subsection{Matched output-sampling comparison}
\label{app:rb}
At a fixed threshold $K$, the CRPS contribution has gradient
\begin{equation}
 g_K=2\frac{\Delta_K}{\pi(K)}
 \bigl(F_\theta(K\mid x)-h_K(r)\bigr)\nabla_\theta F_\theta(K\mid x).
\label{eq:directgrad}
\end{equation}
Boundary marginalization integrates out the randomness of sampled model outputs in the following matched estimator (Proposition~\ref{prop:rb}).

Fix $x,y,\theta$, sample a threshold $K\sim\pi$, and draw two independent complete model outputs $Y_1,Y_2\sim p_\theta$. Set $I_i=\ind\{Y_i\leq v_K\}$, $h=\ind\{y\leq v_K\}$, $w=\Delta_K/\pi(K)$, and $s_i=\nabla_\theta\log p_\theta(Y_i\mid x)$. The score-function estimator of the CRPS gradient at threshold $K$ from two output samples is
\begin{equation}
 G=w(I_1-h)(I_2-h)(s_1+s_2).
\label{eq:samplegrad}
\end{equation}
The directly differentiated gradient is \eqref{eq:directgrad}.

\begin{proposition}[Conditional-expectation variance reduction]
For the matched estimator \eqref{eq:samplegrad}, averaging over model outputs at a fixed threshold gives the direct gradient:
\begin{equation}
 \E[G\mid K,x,y]=g_K.
\end{equation}
With finite second moments, direct differentiation has no greater gradient variance in any parameter direction:
\begin{equation}
 \Cov(G)-\Cov(g_K)=\E_K[\Cov(G\mid K)]\succeq0.
\label{eq:rb}
\end{equation}
\label{prop:rb}
\end{proposition}

\begin{proof}
Condition on a threshold $K$. The two model outputs are independent, and
\[
 \E(I_i-h)=F_\theta(K)-h,\qquad
 \E[(I_i-h)s_i]=\nabla_\theta F_\theta(K),
\]
because $\E s_i=0$ and $\E[I_i s_i]=\nabla_\theta P(Y_i\leq v_K)$. Therefore
\begin{align*}
 \E[G\mid K]
 &=w\E[(I_1-h)s_1]\E[I_2-h]
   +w\E[I_1-h]\E[(I_2-h)s_2]\\
 &=2w(F_\theta(K)-h)\nabla_\theta F_\theta(K)=g_K.
\end{align*}
The law of total covariance gives
\[
 \Cov(G)=\Cov(\E[G\mid K])+\E[\Cov(G\mid K)],
\]
which proves \eqref{eq:rb}. Every conditional covariance matrix is positive semidefinite.

\end{proof}

\section{Implementation Details}
\label{app:impl}
\subsection{Numeral paths and event likelihoods}
\paragraph{Digit tokens and alignment.}
Numerals are constructed from the fixed token IDs representing digits $0,\ldots,b-1$, ordered by digit value, independently of multi-character number tokens. With $n$ prompt tokens, the first $d-1$ path digits follow the prompt, and causal logits are read at zero-based positions $n-1,\ldots,n+d-2$. Padding offsets are handled before gathering logits. We project the required hidden states onto the ten existing vocabulary rows in FP32 with autocast disabled locally, then apply log-softmax over the digit tokens. The backbone uses bfloat16 (BF16), while verification uses 64-bit floating-point (FP64) precision. Each path repeats the prompt and uses ordinary causal attention. For $q$ paths, this layout processes $q(n+d-1)$ input tokens per observation, compared with $n+d-1$ for sequence CE.

\paragraph{One coherent decoder distribution.}
Dropout is disabled in the probability computation while automatic differentiation remains enabled, so paths that share a prefix describe the same decoder distribution. Trainable parameters continue to receive gradients in evaluation mode.

\paragraph{Log-domain operations.}
Log prefix masses are cumulative sums of digit log probabilities. Grouped sibling masses and interval unions use log-sum-exp. Empty digit ranges are skipped, singleton and full-support events have explicit branches, and no probability floor or imputed target is introduced into the objective. Non-finite loss values trigger a diagnostic that records the support, interval, and boundary logits.

\paragraph{Endpoint conversion.}
Measurement intervals are converted to grid indices with decimal or rational arithmetic, preserving endpoint conventions and reporting precision. Empty events are rejected.

\subsection{CRPS estimator details}
\label{app:crpsimpl}
\label{app:crpsextension}
Both sequence NLL and CRPS are strictly proper scores of the complete constrained numeral distribution \citep{gneiting2007}. Sequence NLL sums the autoregressive digit log losses. CRPS weights CDF errors by numerical distance, so adjacent values such as $0999$ and $1000$ remain close despite differing at every digit. The following details specify stable residual evaluation and the stratified estimator.

\paragraph{Stable residual evaluation.}
We compute both $\log F$ and $\log S$ directly from their prefix decompositions and evaluate the squared residual as
\begin{equation}
 \bigl(F_\theta(k)-h_k\bigr)^2=
 \begin{cases}
  \exp(2\log F_\theta(k)),&h_k=0,\\
  \exp(2\log S_\theta(k)),&h_k=1,
 \end{cases}
\label{eq:stablecrps}
\end{equation}
which avoids subtracting a CDF that has rounded to one. All accumulation is performed in at least FP32.

\paragraph{Stratified sampling.}
A stratified design partitions thresholds into fixed contiguous strata $\mathcal S_g$, sets $W_g=\sum_{k\in\mathcal S_g}\Delta_k$, and samples $m_g$ thresholds with probability $\Delta_k/W_g$ inside each stratum. The estimator
\begin{equation}
 \sum_g\frac{W_g}{m_g}\sum_{j=1}^{m_g}
 \bigl[F_\theta(K_{gj})-h_{K_{gj}}\bigr]^2
\label{eq:strata}
\end{equation}
is unbiased and samples every stratum. We split thresholds into two contiguous strata of approximately equal total numerical width and draw one threshold from each.

\begin{algorithm}[!htbp]
\caption{Threshold-sampled CRPS update}
\label{alg:crps}
\begin{enumerate}[leftmargin=*, topsep=2pt, itemsep=2pt]
\item For each exact label, draw $m$ thresholds from a fixed sampling distribution that assigns positive probability to every threshold, or use a stratified design.
\item Teacher-force the threshold paths and compute each log CDF and log survival probability.
\item Evaluate \eqref{eq:stablecrps} and combine terms with the sampling weights in \eqref{eq:samplecrps} or \eqref{eq:strata}.
\item Backpropagate through the threshold paths and update the same trainable parameters.
\end{enumerate}
\end{algorithm}

\section{Verification of the Probability Computation}
\label{app:verify}
The probability computation is verified in FP64 on a three-digit decimal tree with 1,000 outcomes and 70 parameters shared across prefix logits. Checks compare every CDF and 87 interval log probabilities with exhaustive enumeration, test event and CRPS gradients and finite differences, and enumerate the matched score-function expectation on an eight-leaf tree (Table~\ref{tab:verification}). Integration checks cover decoding from digit token IDs, causal logit alignment, shared-prefix consistency, endpoint conversion across carry and rounding boundaries, full-support zero loss and gradient, empty-event handling, and boundary versus enumerated losses and gradients in small causal models. In an FP32 singleton stress test with log mass $-120$, CDF subtraction returns zero while boundary evaluation returns the log mass directly. The minimum eigenvalue of the matched conditional covariance is $-4.41\times10^{-19}$, consistent with roundoff around a positive-semidefinite matrix. 

\begin{table}[!htbp]
\centering\small
\begin{tabular}{@{}lc@{}}
\toprule
Verification quantity & Maximum absolute discrepancy \\
\midrule
CDF versus enumeration & $6.66\times10^{-16}$ \\
Interval log probability versus enumeration & $8.88\times10^{-16}$ \\
Event gradient versus enumeration & $9.99\times10^{-16}$ \\
Event gradient versus finite differences & $2.17\times10^{-11}$ \\
CRPS objective versus threshold expectation & $5.68\times10^{-14}$ \\
CRPS gradient versus threshold expectation & $1.28\times10^{-13}$ \\
Matched score-function mean versus direct gradient & $6.94\times10^{-17}$ \\
\bottomrule
\end{tabular}
\caption{Checks of the probability computation and its derivatives in FP64 on CPU.}
\label{tab:verification}
\end{table}

\section{Experimental Details}
\label{app:exp}
\paragraph{Controlled family.}
The reference family is a truncated mixture on $[-2,2]$,
\begin{align}
 \widetilde p(y\mid x)&=w(x)\mathcal N(y;\mu_1(x),\sigma_1(x)^2)
 +(1-w(x))\mathcal N(y;\mu_2(x),\sigma_2(x)^2),\\
 w(x)&=\operatorname{sigmoid}(3x),\quad
 \mu_1(x)=0.6\sin(\pi x)-0.5,\quad
 \mu_2(x)=0.6\sin(\pi x)+0.5,\nonumber\\
 \sigma_1(x)&=0.06+0.12(x+1)/2,\qquad
 \sigma_2(x)=0.08+0.10(1-x)/2.\nonumber
\end{align}
The density is conditioned on $[-2,2]$, exact bin masses are computed, and the grid-valued target is sampled. The allocation is 8,000/1,000/2,000 train/validation/test inputs.

\paragraph{Reporting mechanisms.}
For the controlled experiment, the reporting mode is drawn independently of the sampled outcome. The nominal precise-label fraction $\rho$ is the probability of reporting the precise grid value. Otherwise, a rounding report or a threshold report is chosen with equal probability. Rounding widths are drawn uniformly from $\{0.2,0.4\}$ with a fixed origin and half-open convention. A threshold $c$ is drawn uniformly from $[-1.5,1.5]$ and the report retains only $Y\leq c$ or $Y>c$, whichever occurred. These mechanisms satisfy \eqref{eq:coarsening}. The same latent samples and reporting random numbers are reused across methods and fractions, with nested precise-label masks for $\rho\in\{0,0.1,0.2,0.5,1\}$. Thus $\rho=0$ retains rounding and threshold reports throughout training and validation. The mixture and histogram heads are evaluated at $\rho\in\{0,0.1,0.2,0.5\}$. The three decoder objectives are compared at $\rho\in\{0.1,0.2,0.5\}$ and share one reference fit per seed at $\rho=1$.

For KBSS, each synthetic budget is drawn independently from the empirical distribution of training grid indices: with probability $0.9$, a sampled index defines a budget at the midpoint above it on the logarithmic grid, and with probability $0.1$, no timeout is imposed. This allows uncensored observations throughout the support and gives an approximate censoring rate of $0.45$ when ties are rare. One censoring configuration is fixed before comparison, and its realized censoring fraction is reported. A separate evaluation budget is fixed from the training data before test evaluation.

\paragraph{Molecular cohort and provenance.}
The hERG cohort uses ChEMBL 37 target \texttt{CHEMBL240}, IC50 measurements in nanomolar (nM) units, positive finite \texttt{standard\_value}, and \texttt{standard\_relation} in $\{=,<,\leq,>,\geq\}$ \citep{chembl37}. We require direct single-protein assignment (assay confidence score 9), retain records with no validity flag or with \texttt{Manually validated} status, and exclude flagged potential duplicates \citep{chemblfaq}. The text forms \texttt{<=}/\texttt{>=} and mathematical-symbol equivalents are normalized. The split manifest records activity, molecule, assay, and document IDs, the release, query, filters, and extraction checksum. The saved manifest records split-specific equality and censoring counts, bound directions, and threshold distributions.

\paragraph{Molecular event conversion.}
Let $C$ be IC50 in nM and $Y=g(C)=9-\log_{10}C$. Then $C>c$ becomes $Y<g(c)$, $C\geq c$ becomes $Y\leq g(c)$, and the other inequalities reverse analogously. We transform the standardized concentration and relation directly: ChEMBL's \texttt{pchembl\_value} is computed for equality records and is not required for inclusion \citep{chemblfaq}. The declared support is $[0,14]$ pIC50 with $10^4$ grid values. Equality records use the nearest grid value with half-up ties. Rounding intervals are used only when reporting precision is documented. Out-of-support equalities and empty or full-support bounds are excluded with recorded counts. Independent measurements remain separate. Parent-molecule and nonchiral Murcko-scaffold groups are assigned by a deterministic allocation that does not use activity labels, targeting 80/10/10 train/validation/test proportions. Prompts contain the parent molecule's simplified molecular input line entry system (SMILES) string, structured assay type, the target name, and the numerical mapping.

\paragraph{Code data and common supports.}
We use the pinned Code-Regression revision and the KBSS latency field in milliseconds. Formatting, name, and literal variants share an abstract syntax tree (AST) template fingerprint of the code. Unparseable inputs use whitespace-normalized text. A seeded group shuffle reserves the largest group for training and targets 80/10/10 row proportions without splitting groups. Code exceeding 2,048 tokenizer tokens is excluded without truncation. A log-spaced grid is fixed from training extrema with a margin of 0.5 in $\log_{10}(\mathrm{ms})$. The Qwen3 comparison reuses the saved dataset and prompt strings, tokenized by its own pinned tokenizer. Split counts, support, and timeout assignments are unchanged.

\paragraph{Cohorts and reporting.}
Table~\ref{tab:cohortcounts} reports the realized report counts by split. Left bounds report that the target lies below a threshold (the upper bounds of Table~\ref{tab:events}), and right bounds report that it lies above one (the lower bounds).

\begin{table}[!htbp]
\centering\small
\begin{tabular}{@{}llrrrr@{}}
\toprule
Dataset & Split & Exact & Rounded & Left bounds & Right bounds \\
\midrule
Controlled ($\rho=0$) & Train & 0 & 3,916 & 2,091 & 1,993 \\
& Validation & 0 & 514 & 250 & 236 \\
& Test & 0 & 999 & 519 & 482 \\
\addlinespace[3pt]
Controlled ($\rho=0.1$) & Train & 853 & 3,482 & 1,877 & 1,788 \\
& Validation & 98 & 470 & 226 & 206 \\
& Test & 183 & 906 & 462 & 449 \\
\addlinespace[3pt]
Controlled ($\rho=0.2$) & Train & 1,654 & 3,095 & 1,667 & 1,584 \\
& Validation & 184 & 424 & 208 & 184 \\
& Test & 403 & 796 & 398 & 403 \\
\midrule
hERG & Train & 8,345 & 0 & 3,288 & 247 \\
& Validation & 966 & 0 & 495 & 24 \\
& Test & 948 & 0 & 518 & 19 \\
\midrule
KBSS & Train & 3,263 & 0 & 0 & 2,686 \\
& Validation & 397 & 0 & 0 & 347 \\
& Test & 744 & 0 & 0 & 0 \\
\bottomrule
\end{tabular}
\caption{Realized report counts for the fixed controlled-data seed and real-data splits. The precise-label probability $\rho$ is nominal: $\rho=0.1$ yields 853 precise training reports out of 8,000. Controlled-test latent outcomes and the true conditional distribution are retained separately for scoring. Every KBSS test outcome remains exact. hERG bounds are expressed after the pIC50 transformation.}
\label{tab:cohortcounts}
\end{table}

The hERG extraction retains 14,850 of 17,099 candidate activities. Exclusions are 1,339 potential duplicates, 587 records below confidence 9, 316 validity flags, five unsupported relations, and two missing structures. The split has 3,860/1,485/1,485 parent/scaffold groups, and every validation and test group contains one report.

KBSS retains 7,437 of 12,615 latency records after the code length limit of 2,048 tokens, excluding 5,178 longer programs without truncation. The training, validation, and test splits contain 2,578/347/349 template groups, respectively. The largest groups contain 1.56\%, 2.02\%, and 2.02\% of the examples in the training, validation, and test splits, respectively. Median full prompt lengths in the training, validation, and test splits are 1,200, 1,209, and 1,326 tokens, respectively, with maxima of 2,140/2,136/2,140. The log-grid support is $[-2.40812210037,2.19253743808]$ in $\log_{10}(\mathrm{ms})$, determined from training extrema with margin 0.5. There are no out-of-support outcomes in any split. Timeouts follow the reporting rule described above. The evaluation budget is $0.03351313227795$ ms, fixed from the training 75th-percentile grid index. The source dataset revision is \texttt{2410c301\allowbreak 318408ea\allowbreak 80eeacd6\allowbreak 41d887d0\allowbreak 39b95dc0}. Group splits and timeout assignments are fixed across methods.

\begin{table}[!htbp]
\centering\small
\begin{tabularx}{\linewidth}{@{}p{.27\linewidth}Y@{}}
\toprule
Method & Supervision and role \\
\midrule
Event likelihood (ours) & Exact probability of each observed interval through the numerical decoder. \\
Complete-case CE & Same decoder trained only on precise observations. \\
Imputed-label CE & Same decoder trained on midpoint representatives for bounded intervals and admissible representatives adjacent to the bounds for censored labels. \\
Event-aware mixture head & Five-component Gaussian mixture on the declared target scale, trained on the same event likelihood using the same backbone. \\
Event-aware histogram head & Fixed 256-bin head using the same backbone and evidence. The cost study also includes a separate dense output with one probability per grid value. \\
\bottomrule
\end{tabularx}
\caption{Core incomplete-label comparison. The distributional heads compare the numeral decoder with other output parameterizations trained on the same incomplete evidence.}
\label{tab:baselines}
\end{table}

\paragraph{Baseline configuration.}
Table~\ref{tab:baselines} summarizes the compared methods. The five-component mixture head predicts means, positive scales, and normalized weights from the final prompt representation. Its input-independent initialization uses equal mixture weights, means evenly spaced across the central 70\% of the declared support, and scale $\max\{0.075(U-L),2\Delta\}$, where $\Delta$ is the grid spacing. This preserves the synthetic initializer on $[-2,2]$ and scales it consistently to pIC50. The histogram uses 256 bins. Imputation uses an admissible endpoint for one-sided bounds and a midpoint index for finite intervals.

\paragraph{Common probability evaluation.}
The primary metrics are hERG NLL over all reports and over the equality subset, and KBSS CRPS on the prespecified physical and logarithmic scales.
For comparable event NLL, continuous heads are integrated over the common grid cells (the interval of values assigned to each grid value) and normalized over the declared support. Histogram mass is allocated by its piecewise-uniform density. Full NLL retains the empirical measurement weights, and separate equality and censoring scores expose the contributions of each report type. For hERG, equality CRPS uses the same grid target across methods, while median MAE uses the transformed reported equality value. For code latency, both CRPS scales and median errors use every original unrounded test value, including any out-of-support outcome, whose CRPS contribution is integrated exactly.

\paragraph{Training configuration.}
Qwen2.5-1.5B is pinned at revision {\small\texttt{8faed761\allowbreak d45a2633\allowbreak 40a05283\allowbreak 43f099c0\allowbreak 5c9a4323}}, and Qwen3-4B-Base at {\small\texttt{906bfd4b\allowbreak 4dc7f14e\allowbreak e4320094\allowbreak d8b41684\allowbreak abff8539}}. Both use rank-16 LoRA with $\alpha=32$ on attention and feed-forward projections. AdamW uses a constant learning rate $10^{-4}$, $(\beta_1,\beta_2)=(0.9,0.999)$, zero weight decay, gradient-norm clipping at 1.0, and effective batch size 32. No search over learning rates or head resolutions is performed. Dropout is disabled while automatic differentiation remains active. Model/tokenizer revisions, digit IDs, source hashes, dataset manifests, support, and conversion rules are saved with each run. Table~\ref{tab:runallocation} lists the fit counts, update budgets, and checkpoint selection rules.

\begin{table}[!htbp]
\centering\small
\begin{tabular}{@{}lcrrl@{}}
\toprule
Study & Backbone & Fits & Max updates & Validation selection \\
\midrule
Controlled incomplete labels & Qwen2.5 & 64 & 750 & Event NLL \\
hERG & Qwen2.5 & 15 & 1,116 & Event NLL \\
KBSS & Qwen2.5 & 15 & 558 & Event NLL \\
KBSS transfer & Qwen3 & 9 & 558 & Event NLL \\
Exact-label scores & Qwen2.5 & 9 & 2,250 & CRPS \\
\bottomrule
\end{tabular}
\caption{The 112 distinct learning fits represented in the reported comparisons. The 64 controlled fits include 60 core fits and four additional event likelihood and complete-case CE fits for seeds 3 and 4 at $\rho=0.2$. The core allocation is 12, 15, 15, 15, and 3 fits at $\rho=0,0.1,0.2,0.5,1$, respectively. The shared full-label reference is counted once per seed.}
\label{tab:runallocation}
\end{table}

The incomplete-label studies run for three epochs, and the exact-label score study runs for nine. Validation includes step zero and epoch/final ends. Synthetic validation occurs every 50 updates, and hERG and both KBSS studies validate every 20 updates. Each checkpoint is selected separately for its method and seed using only observed validation labels. Complete-case CE repeats its precise subset to match the optimizer update budget of the full dataset.

\paragraph{Evaluation budget.}
Synthetic ISE and CRPS use 64 shared stratified threshold queries per input, each query being a gap between adjacent grid values, with a separate integration check comparing 64 and 256 queries at $\rho\in\{0.2,0.5,1\}$. hERG equality CRPS uses 64 fixed, physically weighted stratified queries per equality. KBSS uses 64 shared threshold queries for both CRPS scales. The gap widths on each scale and the exact position of each unrounded test value inside its gap give the physical and logarithmic contributions. Out-of-support CRPS contributions are integrated exactly. The same queries are reused across methods and training seeds. Event NLL and budget probabilities use direct event queries. Reported medians use deterministic CDF bisection. The precision study is separate: 30 configurations, three benchmark replicates each, and five warmup plus 20 measured updates per replicate.

\FloatBarrier
\subsection{Exact-label score protocol}
\label{app:crpsexperiment}
The completed study compares sequence CE, threshold-sampled CRPS, and CDF-BCE on the existing fully precise controlled data: 8,000/1,000/2,000 training/validation/test inputs, support $[-2,2]$, and 10,000 grid values. All three methods use Qwen2.5-1.5B, seeds 0, 1, and 2, effective/micro batch sizes 32/8, and the common 2,250-update budget. Checkpoints are selected by validation CRPS on the observed labels at steps $0,50,\ldots,2250$, using 64 fixed per-input validation queries. Test evaluation uses a separate fixed set of 64 stratified queries per input.

CRPS and CDF-BCE share two contiguous threshold strata, with one fresh threshold per stratum on every presentation of an observation. Thresholds are drawn for the full effective batch before microbatch slicing. The sampled training scores are divided by the fixed support width $W=4$. Reported test CRPS and ISE retain their original units. For independent threshold draws, the unnormalized CDF-BCE estimator is
\begin{equation}
 \widehat\loss_{\mathrm{CDF\text{-}BCE}}
 =-\frac1m\sum_{j=1}^{m}\frac{\Delta_{K_j}}{\pi(K_j)}
 \left[h_{K_j}\log F_\theta(K_j)
 +(1-h_{K_j})\log S_\theta(K_j)\right].
\label{eq:bcecontrol}
\end{equation}
The stratified implementation uses the weights of \eqref{eq:strata} and direct log tails. Expected excess CDF-BCE is a sum of Bernoulli KL divergences weighted by the positive grid spacings $\Delta_k$, so it is also strictly proper on the grid. The comparison holds the decoder and, between the two threshold objectives, the query design fixed. Sequence CE remains the log score of the complete numeral.

\FloatBarrier
\section{Completed Results and Reproducibility}
\label{app:completedresults}

\subsection{Controlled recovery and seed replication}
\begin{figure}[!htbp]
\centering
\paperfigure{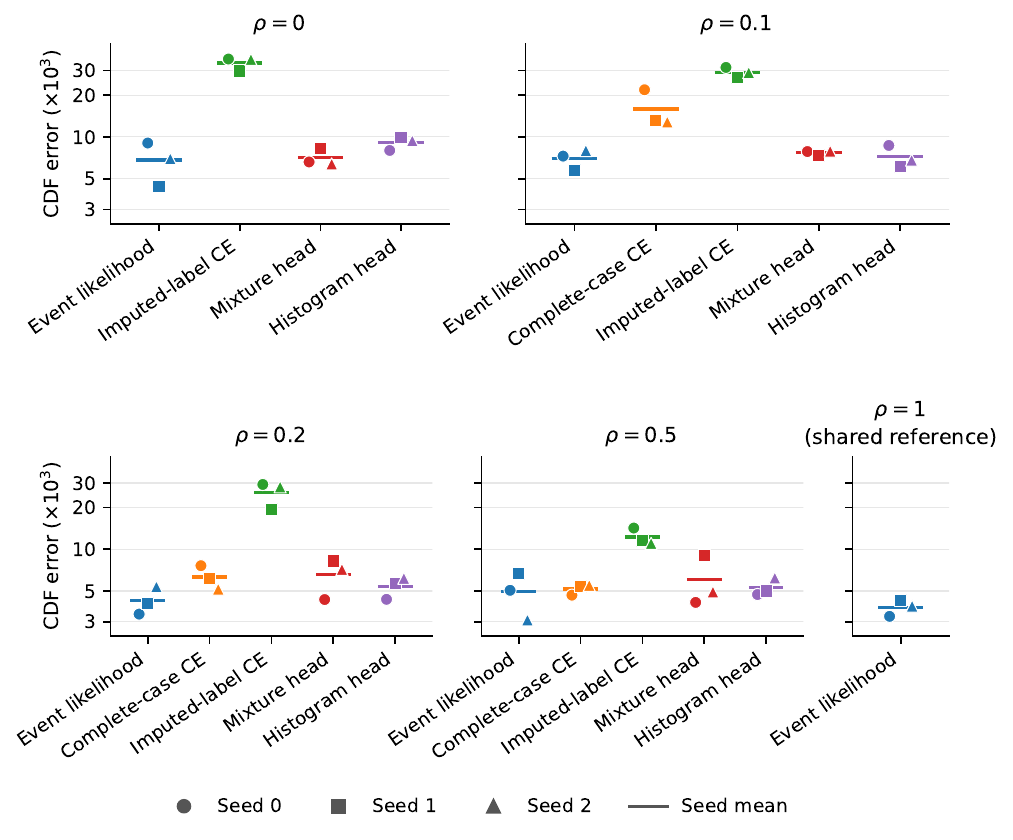}{\linewidth}
\caption{Individual fits behind Figure~\ref{fig:recovery}. Each marker is one training run, identified by its fine-tuning seed, and the bar is the seed mean. All 60 core fits are shown, with the shared full-label reference drawn once per seed. The vertical axis is logarithmic and shows ISE multiplied by $10^3$.}
\label{fig:recoveryseeds}
\end{figure}

\begin{figure}[!htbp]
\centering
\paperfigure{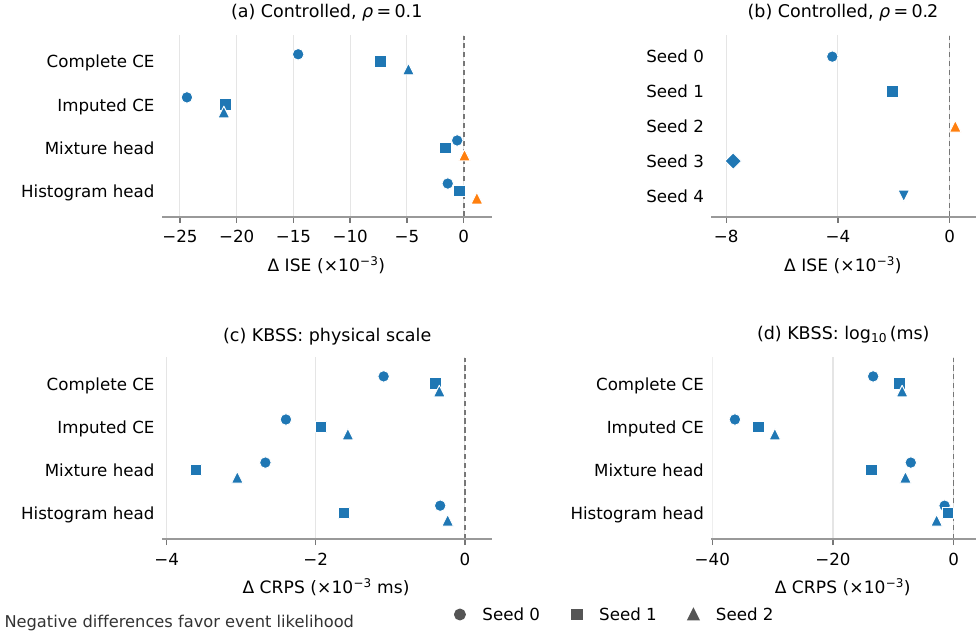}{\linewidth}
\caption{Paired differences across training seeds. (a) ISE differences, event likelihood minus each of the four controls, at $\rho=0.1$ for seeds 0--2. (b) ISE differences, event likelihood minus complete-case CE, at $\rho=0.2$ for all five seeds. (c, d) KBSS CRPS differences, event likelihood minus each of the four controls, for all three Qwen2.5 seeds on physical and logarithmic scales. Negative differences favor event likelihood, and points show individual seed differences.}
\label{fig:pairedeffects}
\end{figure}

Table~\ref{tab:controlledall} gives all 20 three-seed configurations (60 fits). Complete-case CE is unavailable at $\rho=0$, where every training observation is a rounding interval or a one-sided threshold report. The full-label reference at $\rho=1$ is one shared decoder fit per seed. Table~\ref{tab:five} reports the five-seed extension, and Figure~\ref{fig:recoveryseeds} shows the individual fits behind the three-seed results. Figure~\ref{fig:pairedeffects} shows the paired differences between event likelihood and each control at $\rho=0.1$, where event likelihood improves on both CE controls in every seed, together with the five-seed differences at $\rho=0.2$.

\begin{table}[!htbp]
\centering\small
\setlength{\tabcolsep}{3.8pt}
\begin{tabular*}{\linewidth}{@{\extracolsep{\fill}}clcccc@{}}
\toprule
$\rho$ & Method & ISE ($10^{-3}$) & Event NLL & CRPS & Point NLL \\
\midrule
0 & Event likelihood (ours) & $6.801\!\pm\!2.324$ & $0.852\!\pm\!0.009$ & $0.2205\!\pm\!0.0008$ & $8.169\!\pm\!0.067$ \\
0 & Imputed-label CE & $34.121\!\pm\!3.729$ & $1.024\!\pm\!0.017$ & $0.2486\!\pm\!0.0036$ & $9.563\!\pm\!0.086$ \\
0 & Event-aware mixture & $7.099\!\pm\!1.053$ & $0.864\!\pm\!0.010$ & $0.2221\!\pm\!0.0018$ & $7.932\!\pm\!0.006$ \\
0 & Event-aware histogram & $9.085\!\pm\!0.977$ & $0.862\!\pm\!0.022$ & $0.2238\!\pm\!0.0019$ & $8.278\!\pm\!0.026$ \\
\midrule
0.1 & Event likelihood (ours) & $6.993\!\pm\!1.143$ & $1.481\!\pm\!0.005$ & $0.2208\!\pm\!0.0006$ & $7.908\!\pm\!0.018$ \\
0.1 & Complete-case CE & $15.923\!\pm\!5.137$ & $1.544\!\pm\!0.010$ & $0.2322\!\pm\!0.0038$ & $8.153\!\pm\!0.029$ \\
0.1 & Imputed-label CE & $29.156\!\pm\!2.472$ & $1.740\!\pm\!0.011$ & $0.2438\!\pm\!0.0021$ & $9.153\!\pm\!0.052$ \\
0.1 & Event-aware mixture & $7.697\!\pm\!0.303$ & $1.500\!\pm\!0.006$ & $0.2219\!\pm\!0.0025$ & $7.933\!\pm\!0.022$ \\
0.1 & Event-aware histogram & $7.192\!\pm\!1.333$ & $1.522\!\pm\!0.011$ & $0.2211\!\pm\!0.0002$ & $8.315\!\pm\!0.013$ \\
\midrule
0.2 & Event likelihood (ours) & $4.287\!\pm\!1.000$ & $2.263\!\pm\!0.007$ & $0.2193\!\pm\!0.0017$ & $7.900\!\pm\!0.014$ \\
0.2 & Complete-case CE & $6.304\!\pm\!1.230$ & $2.273\!\pm\!0.002$ & $0.2216\!\pm\!0.0012$ & $7.949\!\pm\!0.017$ \\
0.2 & Imputed-label CE & $25.616\!\pm\!5.392$ & $2.569\!\pm\!0.035$ & $0.2420\!\pm\!0.0057$ & $8.869\!\pm\!0.077$ \\
0.2 & Event-aware mixture & $6.574\!\pm\!2.005$ & $2.271\!\pm\!0.008$ & $0.2214\!\pm\!0.0016$ & $7.905\!\pm\!0.028$ \\
0.2 & Event-aware histogram & $5.388\!\pm\!0.935$ & $2.301\!\pm\!0.013$ & $0.2205\!\pm\!0.0024$ & $8.133\!\pm\!0.015$ \\
\midrule
0.5 & Event likelihood (ours) & $4.936\!\pm\!1.780$ & $4.285\!\pm\!0.021$ & $0.2195\!\pm\!0.0015$ & $7.881\!\pm\!0.028$ \\
0.5 & Complete-case CE & $5.182\!\pm\!0.448$ & $4.294\!\pm\!0.012$ & $0.2206\!\pm\!0.0022$ & $7.897\!\pm\!0.013$ \\
0.5 & Imputed-label CE & $12.270\!\pm\!1.706$ & $4.537\!\pm\!0.028$ & $0.2263\!\pm\!0.0012$ & $8.344\!\pm\!0.050$ \\
0.5 & Event-aware mixture & $6.037\!\pm\!2.629$ & $4.285\!\pm\!0.003$ & $0.2207\!\pm\!0.0011$ & $7.879\!\pm\!0.003$ \\
0.5 & Event-aware histogram & $5.312\!\pm\!0.799$ & $4.338\!\pm\!0.028$ & $0.2198\!\pm\!0.0017$ & $8.014\!\pm\!0.031$ \\
\midrule
1 & Shared full-label reference & $3.819\!\pm\!0.499$ & $7.848\!\pm\!0.007$ & $0.2174\!\pm\!0.0012$ & $7.848\!\pm\!0.007$ \\
\bottomrule
\end{tabular*}
\caption{All controlled configurations, mean $\pm$ sample SD over seeds 0, 1, and 2. Event NLL compares methods within the same precise-label fraction, and ISE compares recovery across fractions. CRPS and point NLL score the precise test outcomes. All scores use the fixed 2,000-input test cohort.}
\label{tab:controlledall}
\end{table}

\begin{table}[!htbp]
\centering\small
\begin{tabular}{@{}lccc@{}}
\toprule
Method ($\rho=0.2$) & Seeds & ISE ($10^{-3}$) & Event NLL \\
\midrule
Event likelihood (ours) & 5 & $4.818\!\pm\!1.085$ & $2.258\!\pm\!0.008$ \\
Complete-case CE & 5 & $7.906\!\pm\!2.953$ & $2.288\!\pm\!0.022$ \\
\bottomrule
\end{tabular}
\caption{Five-seed follow-up comparing event likelihood with complete-case CE, extending seeds 0, 1, and 2 with seeds 3 and 4. Entries are means $\pm$ sample SD. The mean ISE difference (event likelihood minus complete-case CE) is $-3.088\times10^{-3}$.}
\label{tab:five}
\end{table}

The paired ISE differences ($10^{-3}$), event likelihood minus complete-case CE, for seeds 0--4 are $-4.201$, $-2.056$, $+0.207$, $-7.750$, and $-1.640$.

Repeated integration checks preserve the direction and magnitude of the gain. On the fixed 256-input subset, the five-seed mean ISE differences are $-2.971$, $-2.960$, $-2.976$, $-2.936$, and $-2.937$ ($\times10^{-3}$) for the primary 64-query draw, two additional 64-query draws, and two 256-query draws. Their range is $4.08\times10^{-5}$ in the unscaled difference.

\paragraph{Conditional CDFs.}
Figure~\ref{fig:conditionalcdf} uses test input \texttt{test\_001617}, selected by the smallest $|x|$ with ties broken by saved ID. The nine fits use their existing checkpoints selected by validation event NLL. We query 512 common grid indices spanning the support, including both endpoints, and compute the true grid CDF through each cell's upper edge, matching the ISE evaluation. The quoted central CDF levels use the shared threshold $v_{4990}\simeq-0.0038$ between the two rises.

\FloatBarrier
\subsection{Supporting hERG and KBSS metrics}
Table~\ref{tab:hergsupport} reports hERG NLL by bound type, equality CRPS, and the MAE of predicted medians. Table~\ref{tab:latencysupport} reports the KBSS budget Brier score and median errors on both scales.

\begin{table}[!htbp]
\centering\small
\setlength{\tabcolsep}{3.8pt}
\begin{tabular*}{\linewidth}{@{\extracolsep{\fill}}lccccc@{}}
\toprule
Method & \shortstack{Censored\\NLL} & \shortstack{Left-bound\\NLL} & \shortstack{Right-bound\\NLL} & \shortstack{Equality\\CRPS} & \shortstack{Equality\\median MAE} \\
\midrule
Event likelihood (ours) & $0.823\!\pm\!0.091$ & $0.801\!\pm\!0.102$ & $1.433\!\pm\!0.296$ & $0.473\!\pm\!0.023$ & $0.655\!\pm\!0.029$ \\
Complete-case CE & $2.076\!\pm\!0.112$ & $2.105\!\pm\!0.119$ & $1.273\!\pm\!0.145$ & $0.436\!\pm\!0.011$ & $0.625\!\pm\!0.019$ \\
Imputed-label CE & $1.356\!\pm\!0.175$ & $1.352\!\pm\!0.189$ & $1.468\!\pm\!0.190$ & $0.454\!\pm\!0.026$ & $0.647\!\pm\!0.039$ \\
Event-aware mixture & $0.850\!\pm\!0.032$ & $0.831\!\pm\!0.035$ & $1.377\!\pm\!0.132$ & $0.480\!\pm\!0.006$ & $0.680\!\pm\!0.008$ \\
Event-aware histogram & $0.963\!\pm\!0.194$ & $0.950\!\pm\!0.203$ & $1.298\!\pm\!0.069$ & $0.513\!\pm\!0.030$ & $0.690\!\pm\!0.030$ \\
\bottomrule
\end{tabular*}
\caption{Supporting hERG scores, mean $\pm$ sample SD over three seeds. There are 518 left bounds and 19 right bounds in the test set. Equality CRPS and the mean absolute error of predicted medians use pIC50 units. Full and equality NLL remain visible in Table~\ref{tab:mainresults}.}
\label{tab:hergsupport}
\end{table}

\begin{table}[!htbp]
\centering\small
\begin{tabular}{@{}lccc@{}}
\toprule
Method & Budget Brier & Median MAE ($10^{-3}$ ms) & Median MAE ($\log_{10}$ ms) \\
\midrule
Event likelihood (ours) & $0.1281\!\pm\!0.0046$ & $9.967\!\pm\!0.226$ & $0.1153\!\pm\!0.0027$ \\
Complete-case CE & $0.1550\!\pm\!0.0082$ & $11.221\!\pm\!0.315$ & $0.1337\!\pm\!0.0052$ \\
Imputed-label CE & $0.1936\!\pm\!0.0020$ & $13.748\!\pm\!0.335$ & $0.1820\!\pm\!0.0079$ \\
Event-aware mixture & $0.1505\!\pm\!0.0026$ & $11.264\!\pm\!0.335$ & $0.1340\!\pm\!0.0075$ \\
Event-aware histogram & $0.1342\!\pm\!0.0036$ & $10.185\!\pm\!0.337$ & $0.1174\!\pm\!0.0053$ \\
\bottomrule
\end{tabular}
\caption{Supporting KBSS metrics on all 744 exact outcomes, mean $\pm$ sample SD over seeds 0, 1, and 2. Brier evaluates strict exceedance of the evaluation budget fixed from the training data, using a direct survival query.}
\label{tab:latencysupport}
\end{table}

\FloatBarrier
\subsection{Cross-backbone supporting scores}
Table~\ref{tab:crosssupport} reports the median errors of the Qwen3-4B-Base study.

\begin{table}[!htbp]
\centering\small
\begin{tabular}{@{}lcc@{}}
\toprule
Method & Median MAE ($10^{-3}$ ms) & Median MAE ($\log_{10}$ ms) \\
\midrule
Event likelihood (ours) & $9.972\!\pm\!0.293$ & $0.1159\!\pm\!0.0038$ \\
Complete-case CE & $11.366\!\pm\!0.494$ & $0.1364\!\pm\!0.0096$ \\
Event-aware histogram & $10.814\!\pm\!0.308$ & $0.1226\!\pm\!0.0033$ \\
\bottomrule
\end{tabular}
\caption{Supporting Qwen3-4B-Base median errors on KBSS, mean $\pm$ sample SD over seeds 0, 1, and 2. Both scores use the same 744 exact outcomes and CDF bisection procedure as the Qwen2.5 study. CRPS and budget Brier scores appear in Table~\ref{tab:crossbackbone}.}
\label{tab:crosssupport}
\end{table}

\FloatBarrier
\subsection{Exact-label score results}
\begin{figure}[!htbp]
\centering
\paperfigure{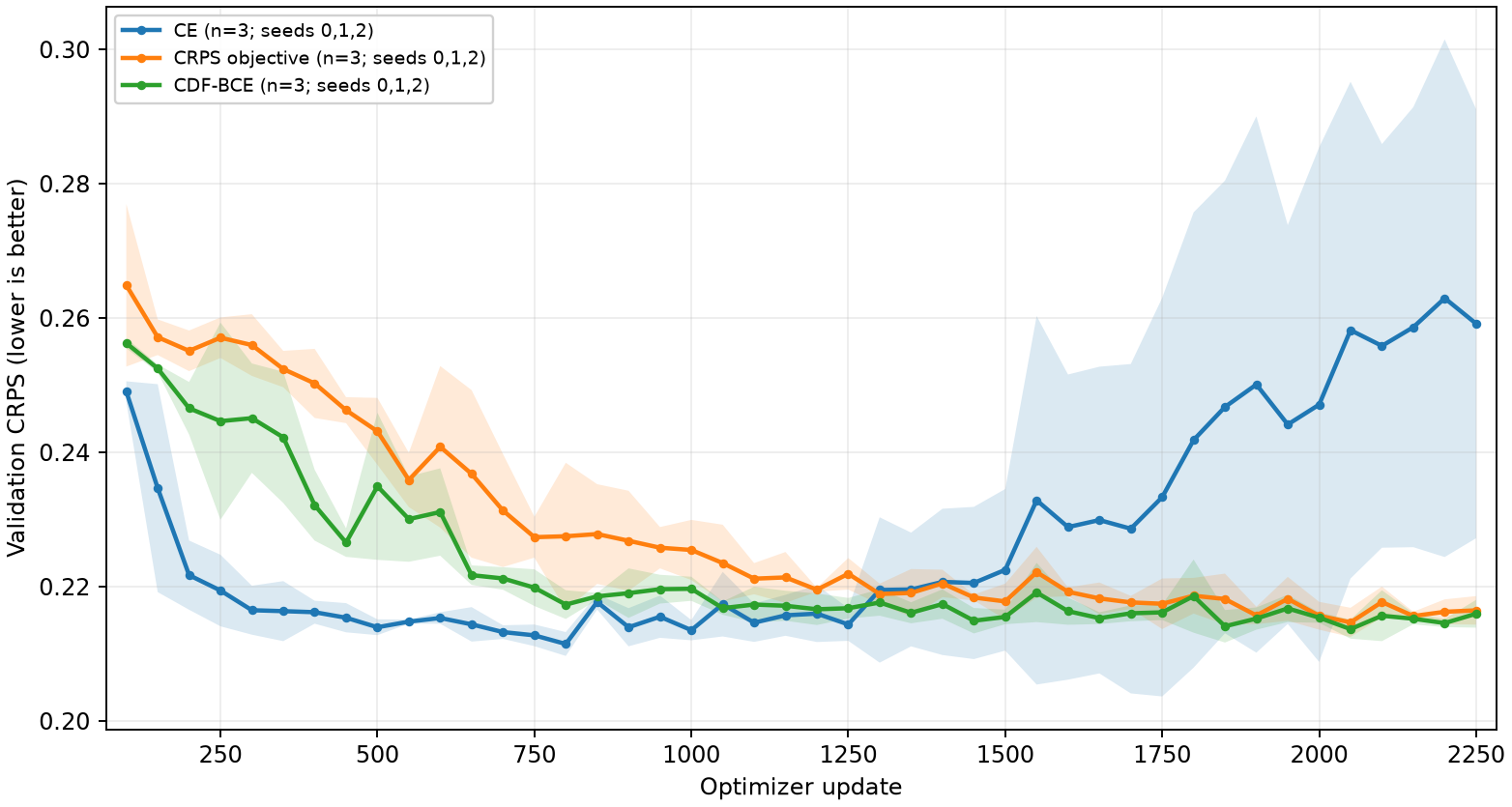}{\linewidth}
\caption{Validation CRPS during the exact-label score study, shown from update 100 through the common 2,250-update budget. Curves are unsmoothed means over seeds 0, 1, and 2, and shading denotes one sample SD across seeds. Every checkpoint uses the same 64 fixed stratified validation queries per input. Test results use the checkpoint with the lowest validation CRPS, selected separately for each method and seed.}
\label{fig:scorecurves}
\end{figure}

Table~\ref{tab:scores} reports the exact-label test scores. The updates selected by validation CRPS for seeds 0, 1, and 2 are $(800,800,750)$ for CE, $(2050,1900,2050)$ for CRPS, and $(2050,1900,1850)$ for CDF-BCE. Figure~\ref{fig:scorecurves} summarizes the validation trajectories. The full-label reference in Table~\ref{tab:controlledall} uses the separate incomplete-label protocol, with 750 updates and checkpoint selection by validation event NLL.

\begin{table}[!htbp]
\centering\small
\begin{tabular}{@{}lccc@{}}
\toprule
Training objective & Test CRPS & ISE ($10^{-3}$) & Point NLL \\
\midrule
Sequence CE & $\mathbf{0.21647}\!\pm\!0.00023$ & $\mathbf{3.399}\!\pm\!0.910$ & $\mathbf{7.842}\!\pm\!0.010$ \\
Threshold CRPS & $0.21974\!\pm\!0.00162$ & $5.777\!\pm\!1.171$ & $8.156\!\pm\!0.062$ \\
Threshold CDF-BCE & $0.22189\!\pm\!0.00202$ & $6.856\!\pm\!1.516$ & $8.180\!\pm\!0.273$ \\
\bottomrule
\end{tabular}
\caption{Exact-label controlled comparison at a common 2,250-update budget. Entries are means $\pm$ sample SD over seeds 0, 1, and 2, and lower is better. Every method selects its checkpoint by validation CRPS. Test CRPS and ISE use 64 common stratified queries on 2,000 test inputs. Point NLL scores the complete precise numeral.}
\label{tab:scores}
\end{table}

\subsection{Complete precision measurements}
\label{app:precisionresults}
The benchmark uses the first 32 input IDs from the controlled training set and four fixed physical intervals: $[-1.73,-0.41]$, $[-0.37,0.83]$, $[0.19,1.71]$, and $[-1.41,1.39]$. All remain interior intervals requiring two boundary paths at every depth. Attended neutral context gives exactly 256 or 1,024 prompt tokens. The effective batch is 32 and microbatch is four observations. Every method uses rank-16/alpha-32 LoRA, BF16 backbone computations, and the production probability arithmetic. All three timing replicates use the same random seed. Each fresh process performs five warmup updates followed by 20 synchronized full optimizer updates. The timer includes gradient clearing, input assembly/transfers, all microbatch forwards, likelihood evaluation, backward, clipping, and AdamW. Replicate case order is randomized, and ratios of update times to those of imputed-label CE are paired within the same replicate and physical GPU.

\begin{figure}[!htbp]
\centering
\paperfigure{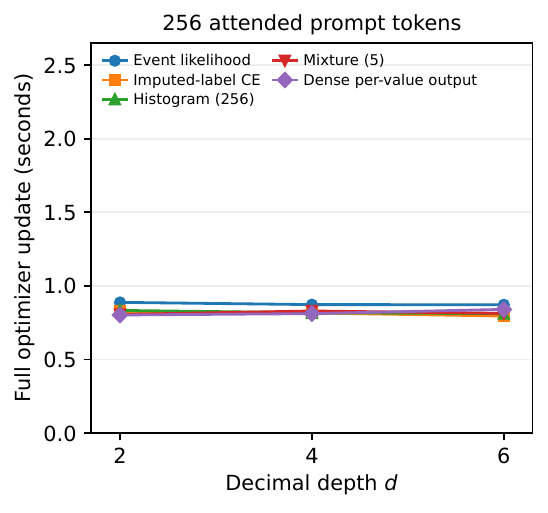}{.48\linewidth}\hfill
\paperfigure{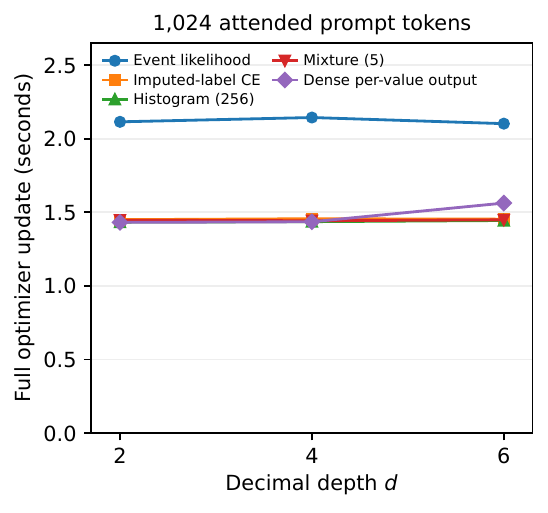}{.48\linewidth}
\caption{Training update time for the same workloads as Figure~\ref{fig:precision}. The event implementation processes two paths per observation, each with its own copy of the prompt, and imputed-label CE and all heads process one prompt. Curves show means of the medians from the three timing replicates.}
\label{fig:precisiontime}
\end{figure}

The dense per-value output predicts one probability per grid value, with $(H+1)10^d$ added output parameters for hidden width $H$. The practical mixture and 256-bin histogram retain fixed parameter counts but store the existing FP64 cell-edge array (about 8 MB at $d=6$). Update times are means of medians from three timing replicates, and memory is the maximum observed peak allocation across replicates. Figure~\ref{fig:precisiontime} shows the update times, and Table~\ref{tab:precisionall} lists every timing and memory measurement.

\begin{table}[!htbp]
\centering\small
\setlength{\tabcolsep}{4pt}
\begin{tabular*}{\linewidth}{@{\extracolsep{\fill}}lrrrrrrr@{}}
\toprule
& & \multicolumn{3}{c}{256 prompt tokens} & \multicolumn{3}{c}{1,024 prompt tokens} \\
\cmidrule(lr){3-5}\cmidrule(l){6-8}
Method & $d$ & Update (s) & GiB & $\times$CE & Update (s) & GiB & $\times$CE \\
\midrule
Event likelihood & 2 & 0.888158 & 11.224 & 1.075 & 2.114650 & 27.458 & 1.458 \\
Event likelihood & 4 & 0.873143 & 11.245 & 1.071 & 2.143870 & 27.071 & 1.472 \\
Event likelihood & 6 & 0.872044 & 11.343 & 1.098 & 2.102270 & 27.008 & 1.446 \\
Imputed-label CE & 2 & 0.826942 & 7.235 & 1.000 & 1.450010 & 19.034 & 1.000 \\
Imputed-label CE & 4 & 0.816379 & 7.245 & 1.000 & 1.456340 & 19.064 & 1.000 \\
Imputed-label CE & 6 & 0.795354 & 7.257 & 1.000 & 1.453370 & 19.095 & 1.000 \\
Histogram (256) & 2 & 0.833546 & 7.234 & 1.009 & 1.433360 & 19.023 & 0.989 \\
Histogram (256) & 4 & 0.817460 & 7.234 & 1.002 & 1.436490 & 19.023 & 0.986 \\
Histogram (256) & 6 & 0.811750 & 7.241 & 1.021 & 1.444380 & 19.031 & 0.994 \\
Mixture (5) & 2 & 0.806047 & 7.228 & 0.976 & 1.446600 & 19.017 & 0.998 \\
Mixture (5) & 4 & 0.829752 & 7.228 & 1.017 & 1.442720 & 19.017 & 0.991 \\
Mixture (5) & 6 & 0.813540 & 7.235 & 1.023 & 1.448850 & 19.025 & 0.997 \\
Dense per-value output & 2 & 0.802479 & 7.230 & 0.973 & 1.431710 & 19.019 & 0.987 \\
Dense per-value output & 4 & 0.811784 & 7.472 & 0.996 & 1.435780 & 19.246 & 0.986 \\
Dense per-value output & 6 & 0.840514 & 35.849 & 1.057 & 1.562770 & 47.513 & 1.075 \\
\bottomrule
\end{tabular*}
\caption{Complete precision scaling measurements. GiB denotes peak allocated memory, and ratios of update times to those of imputed-label CE are computed within each timing replicate and then averaged.}
\label{tab:precisionall}
\end{table}

\clearpage

\bibliographystyle{plain}
\bibliography{references}

\end{document}